\documentclass[lettersize,journal]{IEEEtran}
\usepackage{amsmath,amsfonts}
\usepackage{algorithmic}
\usepackage{algorithm}
\usepackage{array}

\usepackage{textcomp}
\usepackage{float}
\usepackage{url}
\usepackage{verbatim}
\usepackage{cite}
\usepackage{booktabs}
\usepackage{makecell} 
\usepackage{graphicx}
\usepackage{placeins}

\usepackage{subfigure}
\usepackage{bbding}
\usepackage[normalem]{ulem}
\useunder{\uline}{\ul}{}

\begin{document}

\title{Benchmarking Graph Representations and Graph Neural Networks for Multivariate Time Series Classification}

\author{
    Wennuo Yang\textsuperscript{1,*}, Shiling Wu\textsuperscript{1,*}, Yuzhi Zhou\textsuperscript{1}, Cheng Luo\textsuperscript{1}, Xilin He\textsuperscript{1}, \\
    Weicheng Xie\textsuperscript{1,2,3}, Linlin Shen\textsuperscript{1,2,3,†}, Siyang Song\textsuperscript{4,†}\\
    \IEEEauthorblockA{
        \textsuperscript{1}Computer Vision Institute, School of Computer Science \& Software Engineering, Shenzhen University\\
        \textsuperscript{2}Shenzhen Institute of Artificial Intelligence and Robotics for Society\\
        \textsuperscript{3}Guangdong Provincial Key Laboratory of Intelligent Information Processing\\
        \textsuperscript{4}HBUG Lab, University of Exeter\\
        * Equal contribution, † Corresponding authors
    }\\
}



\maketitle

\begin{abstract}
Multivariate Time Series Classification (MTSC) enables the analysis if complex temporal data, and thus serves as a cornerstone in various real-world applications, ranging from healthcare to finance. Since the relationship among variables in MTS usually contain crucial cues, a large number of graph-based MTSC approaches have been proposed, as the graph topology and edges can explicitly represent relationships among variables (channels), where not only various MTS graph representation learning strategies but also different Graph Neural Networks (GNNs) have been explored. Despite such progresses, there is no comprehensive study that fairly benchmarks and investigates the performances of existing widely-used graph representation learning strategies/GNN classifiers in the application of different MTSC tasks. In this paper, we present the first benchmark which systematically investigates the effectiveness of the widely-used three node feature definition strategies, four edge feature learning strategies and five GNN architecture, resulting in 60 different variants for graph-based MTSC. These variants are developed and evaluated with a standardized data pipeline and training/validation/testing strategy on 26 widely-used suspensor MTSC datasets. Our experiments highlight that node features significantly influence MTSC performance, while the visualization of edge features illustrates why adaptive edge learning outperforms other edge feature learning methods. The code of the proposed benchmark is publicly available at \url{https://github.com/CVI-yangwn/Benchmark-GNN-for-Multivariate-Time-Series-Classification}.
\end{abstract}


\section{Introduction}
\noindent Multivariate time series (MTS) represents a series of time stamps containing two or more variables describing different attributes, 
which have been widely investigated to describe various real-world data, including human activities \cite{bagnall2018uea,shokoohi-yekta2017generalizing}, seizures \cite{covert2019temporal,shoeb2009application}, human facial behaviour primitives \cite{song2018human,song2020spectral}, motor imagery \cite{li2021mutualgraphnet,li2023coherence}, etc. Traditional multivariate time series classification (MTSC) solutions can be categorized into distance-based methods \cite{shokoohi-yekta2017generalizing,bagnall2018uea}, shapelet-based methods \cite{ghalwash2012early,li2021shapenet}, dictionary methods \cite{lin2012rotation,dempster2020rocket} and interval-based methods \cite{deng2013time,middlehurst2020canonical}, all of which classify MTS signals based on manually-defined feature learning strategies and traditional machine learning classifiers (e.g., Support Vector Machine (SVM) and Decision Tree (DT)). Consequently, some human uninteratable but informative cues are ignored in such processes.

In contrast to traditional predefined rules approaches, deep learning (DL) models (e.g., Convolution Neural Networks (CNNs), Transformers and Graph Neural Networks (GNNs)) can be optimized to extract task-specific features according to the given data and labels, which have achieved great success in various pattern recognition tasks \cite{wang2017time,zerveas2021transformerbased}.
Consequently, 1D-CNNs have been widely explored for both TSC \cite{wang2017time,ismailfawaz2019deep} and MTSC tasks \cite{yue2022ts2vec,zhao2023classificationoriented,hao2023micos}. Besides, 2D-CNNs also have been applied to extract spatial-temporal features for MTSC tasks \cite{zuo2021smate, pham2023tsem}, which treat each MTS as a 2D feature map. Recent advances in transformers also lead to some transformer MTSC models \cite{zerveas2021transformerbased,cheng2023formertime}. These models take the long-term contexts of time series into account and apply multi-head attention mechanism \cite{vaswani2017attention} to explore different aspects of relevance between variables at different time stamp.

Although existing CNN and transformer-based approaches already achieved promising performances for various MTSC problems, most of them still failed to explicitly model the task-related relationship between each pair of variables, despite that the relationships among variables have been frequently claimed to be crucial for down-stream MTSC tasks (e.g., history speed of roads for traffic forecasting \cite{yu2017spatio}, history electricity consumption for clients for electricity forecasting\cite{wu2020connecting}, EEG signals captured by each electrode for emotion recognition\cite{song2020eeg} and seizure detection\cite{covert2019temporal}). To this end, we notice that the graph is a powerful representation whose topology and edges can explicitly represent relationships among all nodes within it (e.g., variables in a MTS signal). Consequently, some recent MTSC approaches \cite{duan2022multivariate,liu2023todynet} also attempted to represent a MTS signal as a graph, where GNNs were employed to make predictions. There are two main steps for graph-based MTSC: \textbf{graph representation learning for each MTS} and \textbf{GNN-based MTS graph representation classification}. In particular, the graph representation learning stage consists of defining a set of node features, an adjacency matrix defining the graph topology and a set of edge features, where the majority of existing approaches only simply use adjacency matrix with binary/single-value edge features to describe the relationship between each pair of variables \cite{shan2022spatial,wan2022novel,li2020eegbased}. To effectively process such MTSC graph representations, various GNNs \cite{song2020eeg,demir2022eeggat,demir2021eeggnn} also have been employed or developed.

However, existing graph-based MTSC approaches are built on a variety of adjacency matrix, node feature, edge feature learning strategies as well as GNN classifiers, and are evaluated on different datasets/tasks. However, there is no comprehensive study that fairly demonstrates and compares the performances of these graph representation learning strategies/GNN classifiers for the task of MTSC. To bridge this research gap, this paper introduces three main contributions:
\begin{itemize}
    \item We benchmark totally 60 graph-based MTSC systems, including the combinations of 3 widely-used node feature learning strategies, 4 widely-used adjacency matrix learning strategies and 5 GNN classifiers, under a standardized data pipeline and training strategy, on 26 widely-used MTSC datasets (UEA \cite{bagnall2018uea}), i.e., our benchmark provides the first fair evaluation and comparison for graph-based approaches on various MTSC tasks.

    \item  We make the code of all benchmarked models and the standardized pipeline publicly available to facilitate researchers' exploration of new MTSC approaches and apply these methods to more MTSC applications.

    \item Our analysis suggests that node feature learning strategies have a significant impact on the classification performance, and thus this paper provides a practical strategy for selecting appropriate node features based on dataset characteristics. Additionally, the visualization of edge features demonstrates why adaptive edge learning excels as an edge feature learning method.
\end{itemize}

\section{Related work}

\noindent In this section, we first briefly introduce existing Graph construction and typical GNNs in Sec. \ref{subsec:Graph construction}, and then systematically review previous multivariate time series classification approaches (Sec. \ref{subsec:time series classification}).

\subsection{Graph representations and Graph Neural Networks}
\label{subsec:Graph construction}

\noindent \textbf{Graph representation:} Graph has been widely explored to describe various types of real-world data \cite{song2022gratis,dwivedi2020benchmarking}. A regular graph usually consists of three main components: node features, topology and edge features. While each node feature represents the mathematical abstraction of an specific object in the data \cite{hamilton2017inductive}, in the form of a vector or a single value, the topology and edge features of a graph are usually defined by an adjacency matrix. In addition to defining the topology and edge features via the natural connectivity among nodes \cite{yan2018spatial,abbasi2022statistical}, or their similarity \cite{jia2020attentionbased,li2021mutualgraphnet}, some studies propose to deep learn task-specific adjaceny matrix. Ioannidis et al.\cite{ioannidis2019recurrent} treat the relationship between each pair of nodes as a dynamic non-linear process obtained adaptively through a set of learnable weights. Song et al.\cite{song2021uncertain} consider adjacency matrix as a learnable parameter and employ a weighted mask based on Graph Attention Network (GAT) to adaptively capture the importance of each edge (the strength of association between AUs) in a facial graph. Wang et al.\cite{wang2021graphtcn} use a linear transformation to learn the relative spatial relation between a pair of pedestrians and construct a graph to model pedestrians’ interactions. Zhang et al.\cite{zhang2019context} deep learns the affective relationship between each pair of local facial regions as adjacent matrix elements to construct a facial affective graph for emotion recognition. Instead of using single-value edge representations, recent studies \cite{song2022gratis,song2022learning,shao2021personality} have proved that multi-dimensional edge features would provide more informative cues for graph analysis tasks, as they can describe the relationship between node features in a more comprehensive manner. 

\noindent \textbf{Graph Neural Networks:} Existing Graph Neural Networks (GNNs) can be mainly divided into two categories: spectral-based GNN and spatial-based GNN. Spectral-based GNNs approach graph convolutions from the perspective of graph signal processing. They interpret convolution as noise reduction in graph signals by leveraging the spectral decomposition of the normalized graph Laplacian. For instance, Bruna et al. \cite{bruna2013spectral} pioneered this approach by defining the graph Fourier transform based on the eigenvectors of the normalized Laplacian matrix. This method allowed graph signals and learnable filters to be transformed into the frequency domain, convolved, and mapped back to the spatial domain. However, its reliance on eigenvector computation made it computationally expensive and specific to fixed graph structures. To overcome these limitations, Chebyshev Graph Convolutional Network (ChebNet) \cite{defferrard2017convolutional} proposes approximating spectral filters using Chebyshev polynomials, enabling localized feature extraction without depending on graph size. Graph Convolutional Network (GCN) \cite{kipf2017semisupervised} simplified the process further by limiting the polynomial order to 1 and approximating the largest eigenvalue of the Laplacian. This reduced complexity while retaining the ability to aggregate local features effectively, paving the way for scalable GNNs. While spectral-based methods focus on the frequency domain, spatial-based GNNs define graph convolutions directly through neighborhood aggregation. These methods are more computationally efficient and adaptable to various graph structures, leading to their widespread adoption in real-world applications \cite{wu2020comprehensive}. Early works like Neural Network for Graphs (NN4G) \cite{micheli2009neural} aggregated information from node neighborhoods without relying on spectral theory. This simplicity laid the foundation for modern spatial-based methods. Message Passing Neural Network (MPNN) \cite{gilmer2017neural} formalized this approach, introducing update and aggregation functions to govern how nodes exchange information along edges. Building on MPNN, Graph Isomorphism Network (GIN) \cite{xu2018powerful} addressed its inability to differentiate certain graph structures. GIN introduced a learnable parameter to balance the weight of central nodes, enhancing structural discrimination. For large-scale graphs, GraphSAGE \cite{hamilton2017inductive} employed neighborhood sampling to limit computational costs, maintaining efficiency while scaling to dynamic graph structures. Graph Attention Network (GAT) \cite{velickovic2017graph} further advanced spatial-based methods by introducing an attention mechanism. Unlike previous approaches, GAT assigned dynamic importance to neighboring nodes using learnable attention weights. Multi-head attention was added to improve model expressiveness, making GAT a robust choice for heterogeneous graph structures.

\subsection{Multivariate time series classification}
\label{subsec:time series classification}

\noindent \textbf{Non-graph approaches:} Early predefined rules methods for Multivariate Time Series Classification (MTSC) can be broadly categorized into distance-based methods, shapelet-based methods, dictionary-based methods, and interval-based methods \cite{ruiz2021great}. \textbf{Distance-based methods} \cite{gorecki2015multivariate,banko2012correlation} classify time series by measuring similarities between a pair of time-series using predefined distance metrics, most commonly Euclidean Distance (ED) and Dynamic Time Warping (DTW), in combination with classifiers such as K-Nearest Neighbors (KNN) and Support Vector Machine (SVM). Consequently, the accuracy of these approaches heavily relies on precise distance measurements between time series. For example, Banko et al. \cite{banko2012correlation} propose a correlation-based dynamic time warping approach, which aligns time series using DTW as the distance measurement and then applies principal component analysis to compute correlations between multivariate series. \textbf{Shapelet-based methods} emphasize local patterns \cite{ghalwash2012early,karlsson2016generalized,li2021shapenet}, based on the assumption that time series belonging to the same category would contain similar short segments. Ghalwash et al. \cite{ghalwash2012early} propose Multivariate Shapelets Detection which extracts multivariate shapelets (i.e., a vector comprising shapelets from every dimension of the MTS) from different MTS and then compares them. The Generalized Random Forest method introduced by \cite{karlsson2016generalized} performs MTSC by selecting shapelets from MTS using random forest techniques \cite{breiman2001random}. Alternatively, \textbf{dictionary-based methods} \cite{schafer2017multivariate,dempster2020rocket,dempster2021minirocket} represent each time series as a combination of multiple patterns, where each pattern is treated as a `word' in a dictionary. For instance, Schafer et al. \cite{schafer2017multivariate} utilize a sliding window and Symbolic Fourier Approximation to extract unigrams and bigrams, which are then concatenated into a bag-of-words histogram. The Random Convolutional Kernel Transform (ROCKET) \cite{dempster2020rocket} apply convolutional kernels as dictionary sources to generate 10,000 random kernels with varied parameters (e.g., length, weights, and dilation), which then extracts maximum values and the proportion of positive values as input for the classifier. In addition, \textbf{interval-based methods} \cite{deng2013time,middlehurst2020canonical} also segment the given time series into a set of short segments, but focus more on dividing it into non-overlapping chunks and capturing both global and local cues. Samples from the corresponding time chunks across all multivariate dimensions are then used to train classifiers, with the final classification achieved by aggregating results from all chunks. Time Series Forest (TSF) \cite{deng2013time} is a widely-used interval-based approach for univariate time series analysis. It trains a random forest on features derived from randomly selected intervals. The Canonical Interval Forest \cite{middlehurst2020canonical} extends this approach by combining TSF interval features with 22 canonical time-series characteristics \cite{lubba2019catch22}. In addition to hand-crafted methods, \textbf{deep learning (DL)-based MTSC approaches} also have been extensively explored for MTSC in recent years. Zheng et al. \cite{zheng2014time} introduce the Multi-Channel DCNN (MC-DCNN), which processes each dimension of the given MTS independently and thus fails to capture inter-variable relationships. To address this limitation, Zhao et al. \cite{zhao2017convolutional} jointly train multiple filters with channel counts matching the dimensionality of the MTS. Furthermore, Fawaz et al. \cite{ismailfawaz2019deep} conduct a benchmark of several standard deep neural networks, including Multilayer Perceptrons (MLP) \cite{wang2017time}, Fully Convolutional Networks (FCN) \cite{long2015fully}, Residual Networks (ResNet) \cite{he2016deep}, Encoders \cite{serra2018towards}, Multi-scale Convolutional Neural Networks (MCNN) \cite{cui2016multiscale}, Time Le-Net (t-LeNet) \cite{guennec2016data}, Time Warping Invariant Echo State Networks (TWIESN) \cite{tanisaro2016time}, Time-CNN \cite{zhao2017convolutional}, and MC-DCNN, across both univariate and multivariate time series datasets. The results indicate that the ResNet architecture achieved the best performance. Recently, attention mechanisms \cite{vaswani2017attention} have shown effectiveness in enhancing MTSC models. Techniques such as incorporating attention into CNN-based architectures \cite{cheng2020novel,zhang2020tapnet} and leveraging the long-range dependency modeling of Transformer models \cite{zerveas2021transformerbased} have demonstrated promising results. Building on these advancements, Cheng et al. \cite{cheng2023formertime} further optimize Transformer-based models for long time series by implementing temporal reduction techniques.

\noindent \textbf{Graph-based approaches:} Graphs provide a robust representation particularly suited for explicitly modeling the relationships between every pair of variables (or channels) within a MTS. Among existing graph-based MTSC approaches, each variable is commonly represented as a node in the graph, while edges describe the relationships and interactions among these variables. Each \textbf{node feature} in an MTS graph is typically extracted to represent a channel or variable within the MTS. Many approaches directly treat each channel in an MTS as the node feature \cite{li2023coherence,zheng2023spatialtemporal,shen2022multiscale}. Additionally, Differential Entropy (DE), a tool for quantifying information uncertainty, has been widely used as a representation of node features. The process involves transforming the time-series signal into the frequency domain, dividing the frequency spectrum into multiple bands, and calculating the DE for each band separately to serve as the node feature \cite{jiang2021discriminating,li2022eeg,wu2022multistream,zeng2022siamgcan}. Alternatively, Power Spectral Density (PSD) is another important concept in signal analysis. Calculated in the frequency domain, PSD describes the power distribution of a signal across frequencies. By normalizing the power at each frequency point, PSD reflects the proportion of the signal's power at different frequencies \cite{jia2022crgcn,li2020eegbased,klepl2022eegbased}. While the topology of a graph is described by its \textbf{adjacency matrix}, a widely used way for defining MTS graphs involves constructing a fully-connected topology \cite{demir2022eeggat,demir2021eeggnn}, where each node is connected to all others, i.e., all edge features are uniformly set to a single value `1'. However, such uniform edge features fail to capture the differences in relationships between nodes. As a result, recent methods have addressed this limitation by defining MTS graph edge features based on channel relationships from various perspectives. For instance, the Pearson Correlation Coefficient (PCC) \cite{jia2020attentionbased,chen2020epilepsy,hou2022gcnsnet} and Mutual Information \cite{li2021mutualgraphnet,lin2021fatigue} are commonly employed metrics to quantify the strength of associations between every pair of MTS channels/variables (i.e., the similarity between a pair of nodes).
Unlike static edge features, dynamic edge features adapt latent relationships between variates/channels within a MTS during training and inference, and dynamically refine the graph's edge features to better represent the target MTS \cite{wang2021linking, liu2021eegbased, jiang2021discriminating}, e.g., directly updating edge features during back-propagation, which are treated as learnable parameters within the model \cite{song2020eeg}. Additionally, while adjacency matrx-based single-dimensional edge feature definition are predominantly in most graph-based MTSC approaches, multi-dimensional edge features are gaining increasing attention. Multi-dimensional edges allow for the construction of multiple links between nodes, thereby capturing more complex relationships \cite{hou2023audio,luo2022learning,song2024merg}.

\section{The proposed benchmarking framework}

\noindent This paper proposes the first benchmark which provides a rigorous and reproducible evaluation of existing widely-used graph machine learning methods for various MTSC tasks, which conducts a standardized and flexible data loading and model training pipeline. Thus, our benchmark equips future researchers with a set of strong graph-based MTSC baselines. In this section, we first introduce the main pipeline of our benchmarking framework (Sec. \ref{code}) and then provide the details of all the benchmarked components, including three node feature extraction strategies (Sec. \ref{node}), four edge feature learning methods (Sec. \ref{efl}), and four GNNs (Sec. \ref{gnn}) that have been widely employed in previous MTSC approaches. Additionally, we introduce multidimensional edge feature learning strategy on the GAT network, referred to as Multi-dimensional Edge Feature Graph Attention Network (MEGAT) in Sec. \ref{mefl}.

\begin{figure*}[]
    \centering
\includegraphics[width=1\linewidth]{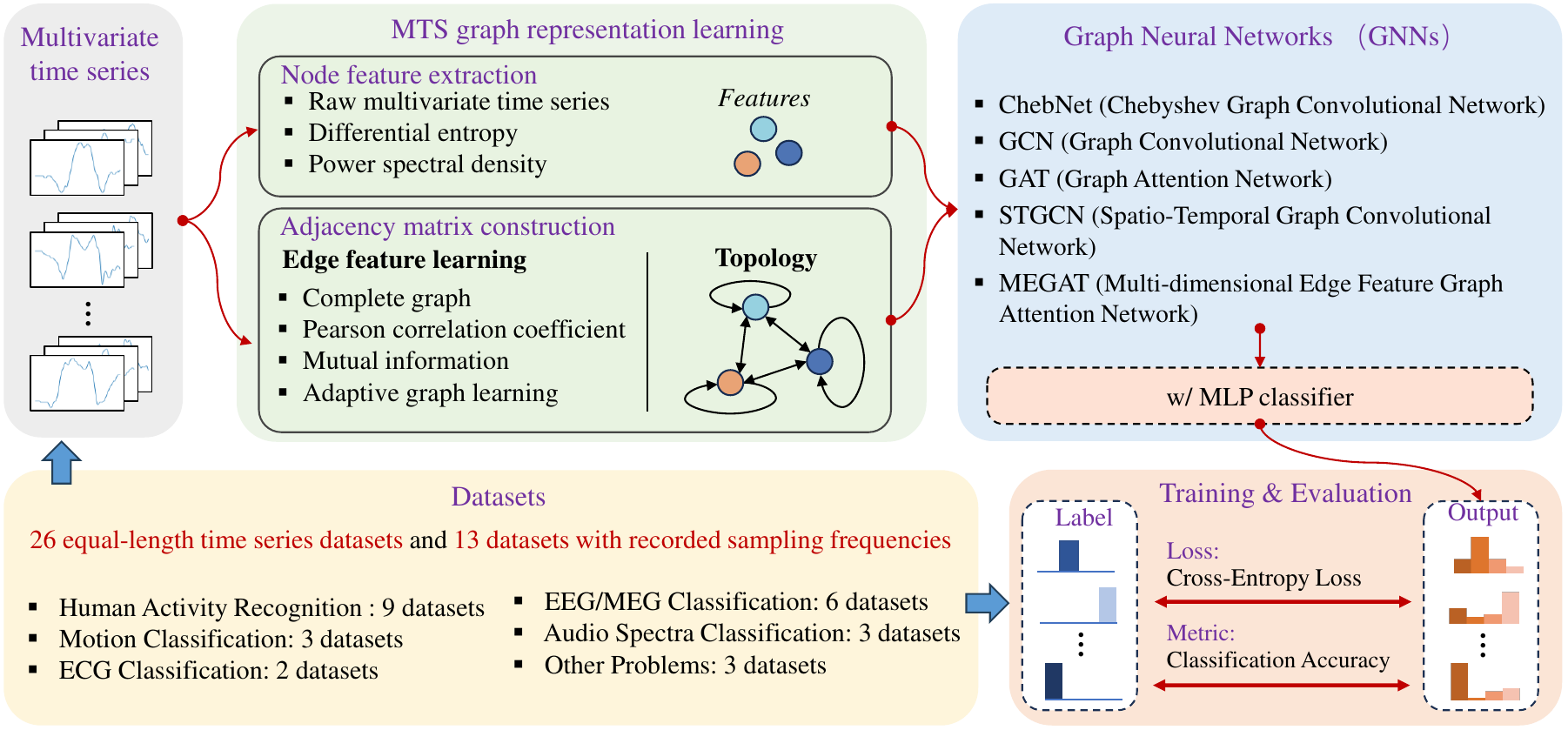}
    \caption{Illustration of the graph neural network-based benchmark framework for multivariate time series classification.}
    \label{fig:benchmark}
\end{figure*}

\subsection{Pipeline and coding infrastructure}
\label{code}

\noindent Our benchmark aims to fairly compare graph-based solutions for the MTSC task. To facilitate a fair comparison, our benchmark emphasizes a unified framework focusing on evaluating the fundamental components of graph-based solutions, including node feature extraction strategies and adjacency matrix definition strategies involved in constructing the input MTS graph representation, as well as the following GNN classifiers. Our benchmarking framework standardizes the entire process for all deep learning models, including data input, model initialization, training, validation, evaluation, and the coding platform/libraries, as detailed in Sec.\ref{sec:training}. Here, we follow a common assumption that have been widely employed by previous MTSC studies \cite{joshi2020transformers,velivckovic2023everything,piao2024garnn,luoknowledge} to fully connect all nodes (i.e., complete graph), as in practice, when the number of node features is small (e.g., typical MTSC), fully-connected topology have been frequently employed \cite{shen2019graph,velivckovic2023everything}. Instead, our benchmark focuses on different strategies in learning edge features. Our benchmark pipeline starts with the extraction of node features while simultaneously constructing an adjacency matrix to capture relationships between nodes, exploring various edge feature methods during this process. These components are then form a MTS graph representation and fed into a GNN to learn a task-related MTS graph representation by leveraging both the node features and the structural information encoded in the adjacency matrix. This graph representation is subsequently processed by a multilayer perceptron (MLP) classifier to produce the final classification results. The entire pipeline is trained using the cross-entropy loss function, with classification accuracy serving as the evaluation metric. The experiments differ only in their fundamental components, designed to evaluate the effectiveness of various methods. Fig. \ref{fig:benchmark} illustrates the the streamlined workflow of our benchmark framework.

\subsection{MTS graph representation learning}
\noindent A large number of multivariate time series classification (MTSC) approaches have been built on graph in the literature. In this paper, we propose the following inclusion criteria to benchmark the most representative and commonly-used graph representation learning and GNN solutions for MTSC task, including:
\begin{itemize}
    \item Three most commonly used node feature extraction methods: raw series \cite{jia2020attentionbased,chen2021interactionaware,shen2022multiscale}, differential entropy \cite{liu2021eegbased,li2022eeg,wu2022multistream,zeng2022siamgcan}, and power spectral density \cite{li2020eegbased,jia2022crgcn,klepl2022eegbased}. 

    \item Four types of most commonly used edge feature learning strategies: Complete graph\cite{sun2017complete,maurya2022simplifying,watkins2023generating}, Pearson Correlation Coefficient \cite{chen2020epilepsy,lu2022pearnet,vijayvargiya2023pc}, mutual information \cite{di2020mutual,li2021mutualgraphnet,liao2022gsaml}, and adaptive edge learning \cite{jia2021graphsleepnet,zhang2022adapgl,xu2023dynamic}.

    \item Three commonly used GNN predictors: Chebyshev Graph Convolutional Network (ChebNet) \cite{defferrard2016convolutional,yan2021spatial,zhang2022deeppn}, Graph Convolutional Network (GCN) \cite{kipf2017semisupervised,zhang2019graph,chen2020simple}, and Spatio-Temporal Graph Convolutional Network (STGCN) \cite{yu2017spatio,covert2019temporal,ghosh2020stacked}, as well as Graph Attention Network (GAT) \cite{velickovic2017graph,wang2019kgat,brynte2024learning} which is not commonly used in MTSC, but has shown excellent performances in various other applications.

    \item A multi-dimensional edge feature learning module \cite{song2022gratis} that have demonstrated the potential to enhance model performance across different types of neighborhoods. \\

\end{itemize}

\subsubsection{Node feature extraction}
\label{node}

Given a MTS $\mathbf{X} = [\mathbf{x}_{1}, \mathbf{x}_{2}, \cdots, \mathbf{x}_{M} ]^T$ that contains $M$ variables recorded at $N$ consecutive time stamps ($\mathbf{x}_{m} = [\mathbf{x}_{m}^{1}, \mathbf{x}_{m}^{2}, \cdots, \mathbf{x}_{m}^{N}]$), we benchmark three widely-used node features as follows: \\

\noindent \textbf{Raw multivariates time series:} Given an MTS $\mathbf{X}$, each raw univariate time series $\mathbf{x_m}$ is directly treated as a node, which retains all original information. \\

\noindent \textbf{Differential entropy:} Differential entropy, as node feature, combines information theory and graph structure that aims to capture node properties by quantifying the `information uncertainty' of the nodes. The DE of a univariate series $\mathbf{x_m}$ with probability density function $P_{x_m}$ can be computed as:
\begin{equation}
\begin{split}
    \text{DE}(x_m) &= -\int_{x_m}^{}P({x_m}) \cdot \log_{2}{P({x_m})}d{x_m}
\end{split}
\end{equation}
In practice, we calculate the DE of each time series by dividing it into 5 frequency bands, forming a comprehensive DE feature for each node. \\

\noindent \textbf{Power spectral density:} Power Spectral Density (PSD) provides a way to quantify the frequency content of a signal. It captures how the signal`s power is distributed across different frequency components, revealing important properties of the signal in the frequency domain. For a univariate series $\mathbf{x}_{m}$, the formula for mathematically calculating the PSD can be obtained as follows:
\begin{equation}
    \text{PSD}_{x_m}(f_k)=\frac{\left|F(f_k)\right|^{2}}{N*f_s}
\end{equation}
where ${F}(f_k)=\sum_{n=0}^{N-1} x_m[n] e^{-j 2 \pi f_k n / N}$ is the discrete Fourier transform. \\  

\noindent Since both DE and PSD require knowledge of the original sampling frequency for their computation in the frequency domain, which is critical but not always available in real-world datasets (including the employed UEA datasets), our analysis of DE and PSD features is limited to the 13 datasets from the UEA archive where the sampling frequency is adequately specified. \\

\subsubsection{Adjacency Matrix-Edge feature learning}
\label{efl}

\noindent Given a set of variables $\mathbf{x}_m$ in a MTS $\mathbf{X}$ ($m = 1, \cdots, M$), this paper benchmarks four widely-used edge feature learning strategies. Specifically, the adjacency matrix is directly derived from the raw MTS data, regardless of the node feature computation method. Here, an identity matrix serves as a placeholder for obtaining learnable edge features to maintain a consistent input format. \\

\noindent \textbf{Complete graph:} In this case, all edge features are set to 1, making it impossible to differentiate the relationships between different nodes. \\

\noindent \textbf{Pearson correlation coefficient:} The Pearson correlation coefficient (PCC) quantifies the linear relationship between two time series by comparing their deviations from their respective means, capturing how the two series vary together over time. The PCC value ranges between -1 and 1, where $\text{PCC}{ij} = 1$ and $\text{PCC}{ij} = -1$ indicate that variables $\mathbf{x}_i$ and $\mathbf{x}j$ are positively and negatively linearly correlated, respectively, while $\text{PCC}{ij} = 0$ indicates no linear correlation, though a non-linear relationship may still exist.
The PCC between two variables $\mathbf{x}_i$ and $\mathbf{x}_j$ can be computed as:
\begin{equation}
    \text{PCC}_{ij} = \frac{\text{E}[(\mathbf{x}_i - \mu_i)(\mathbf{x}_j - \mu_j)]}{\sigma_i\sigma_j}
\end{equation}
where $\mu_i$ and $\mu_j$ are the mean values and $\sigma_i$ and $\sigma_j$ are the standard deviations of $x_i$ and $x_j$ time series, respectively. Since this benchmark focuses on the magnitude of graph weights without considering the direction of correlations (positive or negative), the absolute value of PCC is used. \\

\noindent \textbf{Mutual information:} Mutual Information (MI) quantifies the shared information between two variables, capturing both linear and non-linear dependencies. MI is always non-negative, ranging from 0 to positive infinity. A value of $\text{MI} = 0$ indicates complete independence between the variables, while larger values reflect greater shared information, with no upper limit as the mutual dependence increases. Unlike metrics such as the PCC, MI is particularly effective for analyzing non-linear relationships, making it a powerful alternative in many applications. The MI of two variables \(\mathbf{x}_i\) and \(\mathbf{x}_j\) is defined as:
\begin{equation}
    \text{MI}_{ij} = \sum_{\xi=1}^n \sum_{\zeta=1}^n p\left(x_i^{\xi}, x_j^\zeta\right) \log \frac{p\left(x_i^{\xi}, x_j^\zeta\right)}{p\left(x_i^{\xi}\right) p\left(x_j^\zeta\right)},
\end{equation}
where \(p(\cdot)\) and \(p(\cdot, \cdot)\) denote the marginal and joint probability mass functions, respectively. For discrete variables, the summation is over all possible pairs of values \((x_i^\xi, x_j^\zeta)\). If the variables are continuous, the summation is replaced by an integral over their respective probability density functions. \\

\noindent \textbf{Adaptive edge learning:} Since the aforementioned PCC and MI methods rely on prior knowledge and predefined rules to generate edge features, we additionally benchmark an end-to-end deep learning-based strategy proposed in \cite{jia2021graphsleepnet} to automatically learn and generate the edge features for each MTS`s nodes, where the edge feature \( e_{ij} \) is calculated as:
\begin{equation} \label{eq:diffgraphlearn}
    e_{ij} = g\left(\mathbf{f}_i, \mathbf{f}_j\right) = \frac{\exp \left(\text{ReLU}\left(\mathbf{w}^{T}\left|\mathbf{f}_i - \mathbf{f}_j\right|\right)\right)}{\sum_{n=1}^{N} \exp \left(\text{ReLU}\left(\mathbf{w}^{T}\left|\mathbf{f}_i - \mathbf{f}_n\right|\right)\right)},
\end{equation}
where $e_{ij}$ denotes the edge feature between the \( i \)-th node and \( j \)-th node, \( \mathbf{f}_i \) and \( \mathbf{f}_j \) are feature vectors of nodes \( i \) and \( j \), and \( \mathbf{w} \) is a learnable weight vector. This learning strategy dynamically optimizes the edge feature as part of the model training process, enabling the edge feature to better capture the underlying relationships in the MTS data.

\subsection{Graph Neural Networks}
\label{gnn}

Given a connected graph defined as $\mathcal{G} = (\mathcal{V}, \mathcal{E}, \mathbf{A})$, where $\mathcal{V}$ is a set of nodes with the number of $\left | \mathcal{V} \right |$, $\mathcal{E}$ is a set of edges connecting the nodes defined by a weighted adjacency matrix $\mathbf{A} \in \mathbb{R}^{\left | \mathcal{V} \right | \times \left | \mathcal{V} \right |}$ (each element of $\mathbf{A}$ represents the strength of association between a pair of nodes, i.e., edge features), all benchmarked GNNs utilize both node features and the adjacency matrix as inputs.

\subsubsection{Chebyshev Graph Convolutional Network (ChebNet)}
Chebyshev Graph Convolutional Network (ChebNet) \cite{defferrard2016convolutional} extends convolutional neural networks (CNNs) to graph-structured data by leveraging spectral graph theory. It aims to efficiently perform convolution operations on graphs while avoiding the computational overhead of directly calculating eigenvectors, which is required in traditional spectral graph convolution methods. 

\noindent ChebNet starts by defining the graph Laplacian matrix $\mathbf{L}$ as:
\begin{equation}
    \mathbf{L} = \mathbf{D} - \mathbf{A},
\end{equation}
where $\mathbf{D}$ is the diagonal degree matrix with diagonal elements $D_{ii} = \sum_{j} A_{ij}$, and $\mathbf{A}$ is the adjacency matrix. In practice, the normalized Laplacian is often used to ensure numerical stability:
\begin{equation}
    \mathbf{\tilde{L}} = \mathbf{I}_{\left | \mathcal{V} \right |} - \mathbf{D}^{-\frac{1}{2}}\mathbf{A}\mathbf{D}^{-\frac{1}{2}},
\end{equation}
where $\mathbf{I}_{\left | \mathcal{V} \right |}$ is the identity matrix of size $\left | \mathcal{V} \right |$.
The eigendecomposition of $\mathbf{L}$ transforms it into the frequency domain:
\begin{equation}
    \mathbf{L} = \mathbf{U}\mathbf{\Lambda}\mathbf{U}^T,
\end{equation}
where $\mathbf{U} \in \mathbb{R}^{|\mathcal{V}| \times |\mathcal{V}|}$ contains the eigenvectors of $\mathbf{L}$, and $\mathbf{\Lambda}$ is a diagonal matrix of eigenvalues. The eigendecomposition enables graph signals $\mathbf{x} \in \mathbb{R}^{|\mathcal{V}|}$ to be analyzed in the frequency domain. The graph Fourier transform and its inverse are defined as:
\begin{equation}
    \hat{\mathbf{x}} = \mathbf{U}^T \mathbf{x}, \quad \mathbf{x} = \mathbf{U}\hat{\mathbf{x}}.
\end{equation}
Graph filtering, which applies a filter function $g(\mathbf{\Lambda})$ to the eigenvalues of the Laplacian, is defined as:
\begin{equation}
    \mathbf{y} = g(\mathbf{L})\mathbf{x} = g(\mathbf{U}\mathbf{\Lambda}\mathbf{U}^T)\mathbf{x} = \mathbf{U}g(\mathbf{\Lambda})\mathbf{U}^T\mathbf{x}.
\end{equation}

\noindent However, directly computing $g(\mathbf{\Lambda})$ can be computationally expensive. To address this, ChebNet approximates $g(\mathbf{\Lambda})$ using Chebyshev polynomials of order $K$:
\begin{equation}
    g(\mathbf{\Lambda}) \approx \sum_{k=0}^{K-1} \theta_k T_k(\mathbf{\Lambda}),
\end{equation}
where $\theta_k$ are learnable coefficients, and $T_k(\mathbf{\Lambda})$ is the Chebyshev polynomial of order $k$. This allows the graph convolution operation to be approximated as:
\begin{equation}
\label{chebnet}
    \mathbf{y} = g(\mathbf{L})\mathbf{x} \approx \sum_{k=0}^{K-1} \theta_k T_k(\tilde{\mathbf{L}})\mathbf{x},
\end{equation}
where $\tilde{\mathbf{L}} = \frac{2\mathbf{L}}{\lambda_{\text{max}}} - \mathbf{I}$, and $\lambda_{\text{max}}$ is the largest eigenvalue of $\mathbf{L}$.
A ChebNet layer is defined as:
\begin{equation}
    \mathbf{H}^{(l+1)} = \sigma\left(\sum_{k=0}^{K-1} T_k(\tilde{\mathbf{L}})\mathbf{H}^{(l)}\Theta_k\right),
\end{equation}
where $\mathbf{H}^{(l)} \in \mathbb{R}^{|\mathcal{V}| \times F}$ is the input feature matrix, $\Theta_k \in \mathbb{R}^{F \times F'}$ is a learnable weight matrix, and $\sigma$ is a non-linear activation function (e.g., ReLU).

\subsubsection{Graph Convolutional Network (GCN)}
\label{sec:gcn}

\noindent The Graph Convolutional Network (GCN) \cite{kipf2016semi} was proposed to extend traditional convolutional networks to graph-structured data. Unlike conventional CNNs, GCN efficiently aggregates information from graph nodes and their neighbors, making it well-suited for tasks involving irregular data structures.

\noindent GCN simplifies the spectral graph convolution operation by setting $K = 1$ and approximating $\lambda_{\text{max}} \approx 2$, resulting in a linear function with respect to the graph Laplacian matrix $\mathbf{L}$ \cite{kipf2017semisupervised}:
\begin{equation}
    \mathbf{y} = g(\mathbf{L})\mathbf{x} = \theta_0 \mathbf{x} - \theta_1 \mathbf{D}^{-\frac{1}{2}} \mathbf{A} \mathbf{D}^{-\frac{1}{2}} \mathbf{x},
\end{equation}
where $\theta_0$ and $\theta_1$ are Chebyshev coefficients. By assuming $\theta_0 = -\theta_1$, the convolution is further simplified to:
\begin{equation}
    \mathbf{y} = g(\mathbf{L})\mathbf{x} = \theta (\mathbf{I}_{| \mathcal{V} |} + \mathbf{D}^{-\frac{1}{2}} \mathbf{A} \mathbf{D}^{-\frac{1}{2}}) \mathbf{x}.
\end{equation}
To enhance numerical stability and address issues like gradient vanishing or exploding, GCN applies a renormalization trick \cite{kipf2017semisupervised}, replacing $\mathbf{I}_{| \mathcal{V} |} + \mathbf{D}^{-\frac{1}{2}} \mathbf{A} \mathbf{D}^{-\frac{1}{2}}$ with:
\begin{equation}
    \tilde{\mathbf{D}}^{-\frac{1}{2}} \tilde{\mathbf{A}} \tilde{\mathbf{D}}^{-\frac{1}{2}},
\end{equation}
where $\tilde{\mathbf{A}} = \mathbf{A} + \mathbf{I}_{| \mathcal{V} |}$ and $\tilde{\mathbf{D}}_{ii} = \sum_{j} \tilde{\mathbf{A}}_{ij}$. The addition of $\mathbf{I}_{| \mathcal{V} |}$ introduces self-loops, ensuring all nodes have connections and enhancing gradient flow during backpropagation.

\noindent For multivariate time series (MTS) data, the input shape is $\mathbf{X} \in \mathbb{R}^{| \mathcal{V} | \times N}$, where $N$ represents the series length. In this context, nodes typically represent variables, while edges capture relationships or dependencies between variables. Graph convolution for MTS data is generalized as:
\begin{equation}
    Z = \tilde{\mathbf{D}}^{-\frac{1}{2}} \tilde{\mathbf{A}} \tilde{\mathbf{D}}^{-\frac{1}{2}} \mathbf{X} \mathbf{\Theta} \in \mathbb{R}^{| \mathcal{V} | \times F},
\end{equation}
where $\mathbf{\Theta} \in \mathbb{R}^{N \times F}$ represents filter parameters.

\noindent A GCN layer is defined as:
\begin{equation}
    \mathbf{H}^{(l+1)} = \sigma \left( \tilde{\mathbf{D}}^{-\frac{1}{2}} \tilde{\mathbf{A}} \tilde{\mathbf{D}}^{-\frac{1}{2}} \mathbf{H}^{(l)} \mathbf{W}^{(l+1)} \right),
\end{equation}
where $\mathbf{H}^{(l)}$ is the input feature matrix at the $l^{\text{th}}$ layer, and $\mathbf{W}^{(l+1)}$ is the trainable weight matrix. For the first layer, $\mathbf{H}^{(0)} = \mathbf{X}$.

\subsubsection{Graph Attention Network (GAT)}
\label{sec:gat}

\noindent The Graph Attention Network (GAT) \cite{velivckovic2017graph} introduces attention mechanisms into graph neural networks, addressing the limitation of equal weighting in neighborhood aggregation by dynamically assigning importance to neighbors. This approach allows GAT to learn node interactions without requiring prior knowledge of the graph structure. 

\noindent The input features of nodes $i$ and $j$ at layer $l$ are denoted by $\mathbf{h}^{(l)}_i, \mathbf{h}^{(l)}_j \in \mathbb{R}^{N}$, where $\mathbf{h}^{(0)}_i = \mathbf{x}_i$ and $\mathbf{h}^{(0)}_j = \mathbf{x}_j$ are the initial node features. The attention coefficient, representing the unnormalized importance of node $j$ to node $i$, is computed using a self-attention mechanism with the LeakyReLU activation function :
\begin{equation}
    f^{(l)}_{ij} = \text{LeakyReLU} \left( \mathbf{a}^{T} \left[ \mathbf{W} \mathbf{h}^{(l)}_i \parallel \mathbf{W} \mathbf{h}^{(l)}_j \right] \right),
\end{equation}
where $\parallel$ denotes concatenation, and $\mathbf{a}$ and $\mathbf{W}$ are learnable parameters. Here, $f^{(l)}_{ij}$ measures the raw importance of node $j$ relative to node $i$.
To normalize the attention coefficients across all neighbors $j \in \mathcal{N}_i$ of node $i$, the softmax function is applied:
\begin{equation}
    \alpha^{(l)}_{ij} = \frac{\exp \left( f^{(l)}_{ij} \right)}{\sum_{k \in \mathcal{N}_i} \exp \left( f^{(l)}_{ik} \right)},
\end{equation}
where $\mathcal{N}_i$ is the set of neighbors of node $i$. The normalized coefficient $\alpha^{(l)}_{ij}$ represents the relative importance of node $j$ to node $i$.

\noindent The updated feature of node $i$ at the $(l+1)^{\text{th}}$ layer is computed by aggregating the features of its neighbors, weighted by the attention coefficients:
\begin{equation}
    \mathbf{h}^{(l+1)}_i = \sigma \left( \sum_{j \in \mathcal{N}_i} \alpha^{(l)}_{ij} \mathbf{W}^{(l)} \mathbf{h}^{(l)}_j \right),
\end{equation}
where $\sigma$ is a non-linear activation function, such as ReLU.

\subsubsection{Spatio-Temporal Graph Convolutional Network (STGCN)}

\noindent Multivariate time series (MTS) data exhibit both spatial dependencies (relationships among variables) and temporal dynamics (evolving patterns over time). Spatio-Temporal Graph Convolutional Network (STGCN) \cite{yu2017spatio} is designed to simultaneously model these two aspects, making it particularly effective for capturing complex spatio-temporal relationships in MTS data.

\noindent The 2D input variables $\mathbf{H}^{(0)} = \mathbf{X} \in \mathbb{R}^{|\mathcal{V}| \times N}$, where $|\mathcal{V}|$ is the number of graph nodes and $N$ is the number of time steps, are extended to 3D variables $\mathbf{\mathcal{H}}^{(0)} \in \mathbb{R}^{|\mathcal{V}| \times N \times C_0}$, where $C_0 = 1$ initially represents the feature dimension. To capture temporal dynamics, the temporal convolutional layer uses 1D causal convolution with a kernel of width $K_t$. This convolution is followed by gated linear units (GLU) for non-linear activation. The temporal convolution is defined as:
\begin{equation}
{\mathcal{H}_t}^{(l+1)}=(P^{l+1}+\mathcal{H}^l) \odot \sigma(Q^{l+1}),
\end{equation}
where $P, Q \in \mathbb{R}^{(M-K_t+1) \times C_l}$ are the outputs of the $l+1$ causal convolutional layer, split along the channel dimension. Here, $\mathcal{H}^l$ represents the input to the $l+1$ layer, and residual connections are incorporated to ensure stable gradient flow and enhance learning efficiency.

\noindent To model spatial dependencies, the graph convolutional layer is employed (as detailed in Sec. \ref{sec:gcn}). It captures relationships between variables by leveraging the graph structure defined over the MTS data. An STGCN block integrates two temporal convolutional layers and one spatial convolutional layer. Given the input $\mathbf{\mathcal{H}}^{(l)} \in \mathbb{R}^{|\mathcal{V}| \times T_l \times C_l}$ at block $l$, the output $\mathbf{\mathcal{H}}^{(l+1)} \in \mathbb{R}^{|\mathcal{V}| \times T_{l+1} \times C_{l+1}}$ is computed as:
\begin{equation}
\mathbf{\mathcal{H}}^{(l+1)} = {\text{TConv}_1}^{l+1} \left( {\text{SConv}}^{l+1} \left( {\text{TConv}_0}^{l+1} (\mathbf{\mathcal{H}}^{(l)}) \right) \right),
\end{equation}
where ${\text{TConv}_0}^{l+1}$ and ${\text{TConv}_1}^{l+1}$ are the $l+1$-th temporal convolutional layers, and ${\text{SConv}}^{l+1}$ is the $l+1$-th graph convolutional layer.

\subsubsection{Multi-dimensional Edge Feature Learning Graph Attention Network (MEGAT)}
\label{mefl}

Most existing graph representation learning methods use a single value to represent edge features (Sec. \ref{efl}), which often limits their ability to capture the complex relationships between multivariate time series (MTS) variables represented by node feature vectors. To overcome this limitation, we adopt the multi-dimensional edge feature learning module proposed in \cite{luo2022learning} and use in GAT. This module learns edge feature vectors with a dimensionality matching the number of node features, enabling the model to encode richer global contextual relationships for each edge.

\noindent The multi-dimensional edge feature $\mathbf{e}_{ij}$ is computed in two stages. First, a global series representation $\mathbf{G}$ is generated using a 1D convolutional layer. Next, cross-attention operations are applied to extract node relationships from the global context. For a given node feature vector $\mathbf{X}_i$, the cross-attention operation is defined as:
\begin{equation}
    \text{CA}(\mathbf{X}_i, \mathbf{G}) = \mathbf{X}_i \mathbf{W}_q (\mathbf{G} \mathbf{W}_k)^\top (\mathbf{G} \mathbf{W}_v),
\end{equation}
where $\mathbf{W}_q$, $\mathbf{W}_k$, and $\mathbf{W}_v$ are learnable weight matrices. This operation identifies activation cues from the global series representation $\mathbf{G}$ and extracts features that highlight the relationships between nodes.

\noindent The final multi-dimensional edge feature $\mathbf{e}_{ij}$ is obtained by aggregating features through global average pooling ($\text{GAP}$) applied to the outputs of cross-attention operations:
\begin{equation}
\resizebox{1.0\columnwidth}{!}{
$\mathbf{e}_{ij} = \text{GAP}(\text{CA}(\text{CA}(\mathbf{X}_i, \mathbf{G}), \text{CA}(\mathbf{X}_j, \mathbf{G})), (\text{CA}(\mathbf{X}_j, \mathbf{G}), \text{CA}(\mathbf{X}_i, \mathbf{G}))).$
}
\end{equation}

\begin{table*}[ht]
    \caption{A summary of the 26 datasets in the UEA Multivariate Time Series Classification archive. FS denotes the sampling frequency.}
    \label{tab:uea}
    \centering
    \Huge
    \resizebox{0.85\linewidth}{!}{
    \renewcommand{\arraystretch}{1.0}
        \begin{tabular}{l|ccccccc}
            \hline
            Dataset                   & Type   & Train Cases & Test Cases & Dimensions & Length & Classes & FS      \\ \hline
            ArticularyWordRecognition & MOTION & 275         & 300        & 9          & 144    & 25   &\Checkmark      \\ 
            AtrialFibrillation        & ECG    & 15          & 15         & 2          & 640    & 3       & \Checkmark      \\ 
            BasicMotions              & HAR    & 40          & 40         & 6          & 100    & 4       &\Checkmark      \\ 
            Cricket                   & HAR    & 108         & 72         & 6          & 1197   & 12      &\Checkmark      \\ 
            DuckDuckGeese             & AUDIO  & 50          & 50         & 1345       & 270    & 5       &\XSolidBrush      \\ 
            EigenWorms                & MOTION & 128         & 131        & 6          & 17984  & 5       &\XSolidBrush      \\ 
            Epilepsy                  & HAR    & 137         & 138        & 3          & 206    & 4       &\Checkmark      \\ 
            EthanolConcentration      & OTHER  & 261         & 263        & 3          & 1751   & 4       &\XSolidBrush      \\ 
            ERing                     & HAR    & 30          & 270         & 4          & 65     & 6       &\XSolidBrush      \\ 
            FaceDetection             & MEG    & 5890        & 3524       & 144        & 62     & 2       &\Checkmark      \\ 
            FingerMovements           & EEG    & 316         & 100        & 28         & 50     & 2       &\Checkmark      \\ 
            HandMovementDirection     & EEG    & 160         & 74        & 10         & 400    & 4       &\XSolidBrush      \\ 
            Handwriting               & HAR    & 150         & 850        & 3          & 152    & 26      &\XSolidBrush      \\ 
            Heartbeat                 & AUDIO  & 204         & 205        & 61         & 405    & 2       &\XSolidBrush      \\ 
            Libras                    & HAR    & 180         & 180        & 2          & 45     & 15      &\XSolidBrush      \\ 
            LSST                      & OTHER  & 2459        & 2466       & 6          & 36     & 14      &\XSolidBrush      \\ 
            MotorImagery              & EEG    & 278         & 100        & 64         & 3000   & 2       &\Checkmark      \\ 
            NATOPS                    & HAR    & 180         & 180        & 24         & 51     & 6       &\XSolidBrush      \\ 
            PenDigits                 & MOTION & 7494        & 3498       & 2          & 8      & 10      &\XSolidBrush      \\ 
            PEMS-SF                   & OTHER  & 267         & 173        & 963        & 144    & 7       &\XSolidBrush      \\ 
            Phoneme                   & AUDIO  & 3315        & 3353       & 11         & 217    & 39      &\XSolidBrush      \\ 
            RacketSports              & HAR    & 151         & 152        & 6          & 30     & 4       &\Checkmark      \\ 
            SelfRegulationSCP1        & EEG    & 268         & 293        & 6          & 896    & 2       &\Checkmark      \\ 
            SelfRegulationSCP2        & EEG    & 200         & 180        & 7          & 1152   & 2       &\Checkmark      \\ 
            StandWalkJump             & ECG    & 12          & 15         & 4          & 2500   & 3       &\Checkmark      \\ 
            UWaveGestureLibrary       & HAR    & 120         & 320        & 3          & 315    & 8       &\Checkmark      \\ \hline
        \end{tabular}%
    } 
\end{table*}

\begin{table}[]
    \caption{Different methods of constructing node and edge feature and their corresponding abbreviations}
    \label{table:abb}
    \centering
    \resizebox{1.0\linewidth}{!}{
    \begin{tabular}{ll|ll}
    \hline
    \textbf{Node} & \textbf{Abb.} & \textbf{Edge} & \textbf{Abb.} \\ 
    \hline
    Raw series       & Raw            & Complete graph        & CG            \\ 
    Differential entropy              & DE            & Person correlation coefficient                  & PCC           \\ 
    Power spectral density              & PSD           & Mutual information                   & MI           \\ 
    \multicolumn{2}{c|}{} & Adaptive Edge Learning & AEL            \\ 
    \hline
    \end{tabular}
    }
\end{table}

\subsection{Model setting, training and evaluation}
\label{sec:training}

\begin{table*}[ht]
\caption{Average test accuracy (\%) across 13 datasets with sampling frequency, achieved by the benchmarked approaches and tested on five graph neural networks. Bold values indicate the edge feature learning method that performs best for a specific node feature, while underlined values indicate the node feature extraction method that performs best for a specific edge feature learning method.}
\label{table:resultwithfs}
\resizebox{1.0\linewidth}{!}
{
\begin{tabular}{llll|lll|lll|lll|lll}
\hline
 & \multicolumn{3}{c|}{\textbf{ChebNet}} & \multicolumn{3}{c|}{\textbf{GCN}} & \multicolumn{3}{c|}{\textbf{GAT}} & \multicolumn{3}{c|}{\textbf{MEGAT}} & \multicolumn{3}{c}{\textbf{STGCN}} \\ \hline
 & \multicolumn{1}{c}{Raw} & \multicolumn{1}{c}{DE} & \multicolumn{1}{c|}{PSD} & \multicolumn{1}{c}{Raw} & \multicolumn{1}{c}{DE} & \multicolumn{1}{c|}{PSD} & \multicolumn{1}{c}{Raw} & \multicolumn{1}{c}{DE} & \multicolumn{1}{c|}{PSD} & \multicolumn{1}{c}{Raw} & \multicolumn{1}{c}{DE} & \multicolumn{1}{c|}{PSD} & \multicolumn{1}{c}{Raw} & \multicolumn{1}{c}{DE} & \multicolumn{1}{c}{PSD} \\
 CG & {\ul 66.427} & \textbf{63.812} & 57.617 & 62.763 & {\ul 63.622} & 59.774 & {\ul 68.369} & 58.633 & 57.423 & {\ul 68.089} & 63.471 & 62.533 & {\ul 66.177} & 59.962 & 53.460 \\
PCC & {\ul 65.616} & 62.904 & 58.676 & {\ul 64.687} & 63.319 & \textbf{60.694} & {\ul \textbf{69.067}} & 57.105 & 56.005 & {\ul \textbf{69.476}} & \textbf{64.151} & 62.433 & {\ul 67.197} & 61.221 & 55.021 \\
MI & {\ul \textbf{66.615}} & 62.606 & 58.851 & {\ul 65.991} & 62.929 & 60.031 & {\ul 68.869} & 57.766 & 56.593 & {\ul 69.299} & 63.109 & 62.390 & {\ul \textbf{68.340}} & \textbf{62.762} & 53.980 \\
AEL & {\ul 64.987} & 62.696 & \textbf{62.060} & {\ul \textbf{66.823}} & \textbf{66.435} & 60.056 & {\ul 67.238} & \textbf{59.101} & \textbf{58.667} & {\ul 66.986} & 63.891 & \textbf{62.994} & {\ul 64.534} & 61.722 & \textbf{56.920} \\ 
\hline
\end{tabular}
}
\end{table*}

\noindent \textbf{Model settings:} All GNN networks are set to consist of 3 stacked layers, with a hidden layer size of 128 in each layer and the hidden dimension of the MLP classifier is set to match the number of categories in each dataset. For the ChebNet-specific parameter $K$ is set to 3 and the kernel size of the one-dimensional causal convolution in STGCN is set to 3. The batch size for training is 64, but when GPU memory usage exceeds capacity, we iteratively halve the batch size and reduce the number of hidden layers to fit available resources. \\

\noindent \textbf{Model training:} All GNN models are initialized using Kaiming uniform initialization, while the linear and 1D convolutional layers are initialized via Glorot’s uniform initialization to ensure their balanced weight distribution. These models are trained based on a uniformed strategy: using Stochastic Gradient Descent (SGD) with an initial learning rate of 0.001, which is dynamically reduced by 50\% if the training loss does not improve for 10 consecutive epochs, with a minimum value capped at 1e-6. Training is conducted for 200 epochs. Classification accuracy (ACC) is used as the evaluation metric to select the best model, which is determined by the best performance across three independent experiments using random seeds 42, 152, and 310, to mitigate the impact of random variations. \\

\noindent \textbf{Model evaluation and metrics:} All reported approach evaluation results are achieved on the model achieving the minimum loss during the training stage. Following previous studies \cite{li2021shapenet,zhang2024multivariate}, we use classification accuracy (ACC) as the evaluation metric. ACC is a widely adopted metric for classification tasks, as it intuitively measures the proportion of correctly classified samples. It is defined as:
\begin{equation}
\text{ACC} = \frac{1}{N}\sum_{i=1}^{N} \mathbb{I}(y_i = \hat{y}_i),
\end{equation}
where $N$ is the total number of evaluated time-series samples, $y_i$ is the ground truth label of the $i$-th sample, $\hat{y}_i$ is the predicted label, and $\mathbb{I}(y_i = \hat{y}_i)$ is an indicator function equal to 1 if $y_i = \hat{y}_i$, and 0 otherwise. \\

\noindent \textbf{Benchmarking platform:} All benchmarking are performed on Nvidia GPUs, including Tesla P100 and Tesla A100, to accommodate datasets of varying sizes and memory requirements. The entire benchmarking framework are implemented in PyTorch and executed on the Ubuntu operating system.

\subsection{Evaluation datasets}

The UEA Multivariate Time Series Classification (MTSC) archive \cite{bagnall2018uea} contains 30 datasets, encompassing a broad range of cases, dimensions, and series lengths. These datasets were curated by researchers from the University of East Anglia and the University of California, Riverside. To ensure the validity and consistency of our experiments, we benchmark our approaches on 26 out of the 30 MTSC archive problems, focusing specifically on datasets with time series of equal lengths. Additionally, we use 13 datasets with recorded sampling frequencies, which allow the application of DE and PSD as node features. The details of these datasets are presented in Table \ref{tab:uea}.
According to their areas of application, the datasets are categorized into six groups: 
\begin{itemize}
    \item \textbf{Human Activity Recognition (HAR): 9 datasets} HAR focuses on predicting human activities, such as walking or running, using accelerometer or gyroscope data. Its popularity stems from the simplicity of data collection and wide applicability in wearable technology and health monitoring.
    \item \textbf{Motion Classification: 3 datasets} Motion classification focuses on focuses on specific movement patterns beyond general human activities, such as mouth movements during speech or pen-tip dynamics in handwriting, often used in gesture recognition and fine-motor analysis.
    \item \textbf{Electrocardiogram (ECG) Classification: 2 datasets} ECG classification targets health-related applications, including predicting atrial fibrillation outcomes and activity-based heart monitoring, showcasing the potential of MTS in personalized medicine.
    \item \textbf{Electroencephalography/Magnetoencephalography (ECG/MEG) Classification: 6 datasets} EEG and MEG data collected from brain-computer interface competitions, emphasizing applications in cognitive neuroscience, medical diagnosis, and human-computer interaction.
    \item \textbf{Audio Spectra Classification: 3 datasets} This category transforms audio signals into multivariate time series using spectral features, enabling analysis in tasks like sound classification and acoustic event detection.
    \item \textbf{Other Problems:} three datasets do not fit into the aforementioned categories.
\end{itemize}

\section{Experiments}

\noindent In this section, we first present the experimental results of the benchmarked methods and provide relevant comparisons in Sec. \ref{sec:benchmarking results}. Subsequently, in Sec. \ref{sec:analysis}, we perform an in-depth analysis of the results, offering practical suggestions and visualizations to support the findings.

\subsection{Benchmarking results}
\label{sec:benchmarking results}
In this section, we provide an overall description of the experimental results. Table \ref{table:abb} summarizes the abbreviations for node and edge features used in this benchmark, ensuring clarity and consistency in notation. The experimental outcomes are detailed in Table \ref{table:resultwithfs} (datasets with sampling frequency) and Table \ref{table:resultwithoutfs} (datasets without sampling frequency). Specifically, we analyze the results from four key perspectives:
\begin{itemize}
    \item Node Features: Investigating the impact of different node features.
    \item Edge Features: Comparing the performance of various edge feature types.
    \item Single-Dimensional vs. Multi-Dimensional Edges: Evaluating the benefits of multi-dimensional edge features.
    \item GNN Architecture: Assessing the influence of different Graph Neural Network (GNN) architectures on performance.
\end{itemize}

\textbf{Comparison between node feature extraction:} As shown in Table \ref{table:resultwithfs}, using raw series data as node features achieves the best average performance across datasets and demonstrates notable robustness, which consistently outperforms other methods by a significant margin across various networks. In comparison, DE (from an information theory perspective) and PSD (derived from the power spectrum) extract features by transforming data into the frequency domain. While DE performs better than PSD, both methods fall short of the overall performance achieved by the raw series method.

\begin{table}[]
\caption{Average test accuracy (\%) across 13 datasets without sampling frequency (i.e., using only raw data as node features), achieved by the benchmarked approaches and tested on five graph neural networks. Bold values indicate the edge feature learning method that performs best with raw series as node features.}
\label{table:resultwithoutfs}
\centering
\begin{tabular}{llllll}
\hline
 & \multicolumn{1}{c}{\textbf{ChebNet}} & \multicolumn{1}{c}{\textbf{GCN}} & \multicolumn{1}{c}{\textbf{GAT}} & \multicolumn{1}{c}{\textbf{MEGAT}} & \multicolumn{1}{c}{\textbf{STGCN}} \\ \hline
CG & 46.055 & 47.405 & 45.854 & \textbf{53.581} & 47.996 \\
PCC & 45.438 & 48.390 & 45.335 & 52.536 & 49.137 \\
MI & \textbf{51.620} & 49.024 & 45.531 & 53.255 & 48.493 \\
AEL & 44.654 & \textbf{51.498} & \textbf{46.185} & 52.783 & \textbf{50.520} \\ \hline
\end{tabular}
\end{table}

\begin{table*}[t]
\caption{Average test accuracy (\%) on HAR dataset type achieved by the benchmarked approaches and tested on five graph neural networks. Bold values indicate the edge feature learning method that performs best for a specific node feature, while underlined values indicate the node feature extraction method that performs best for a specific edge feature learning method.}
\label{table:HAR}
\resizebox{1.0\linewidth}{!}{
\begin{tabular}{llll|lll|lll|lll|lll}
\hline
 & \multicolumn{3}{c|}{\textbf{ChebNet}} & \multicolumn{3}{c|}{\textbf{GCN}} & \multicolumn{3}{c|}{\textbf{GAT}} & \multicolumn{3}{c|}{\textbf{MEGAT}} & \multicolumn{3}{c}{\textbf{STGCN}} \\ \hline
 & \multicolumn{1}{c}{Raw} & \multicolumn{1}{c}{DE} & \multicolumn{1}{c|}{PSD} & \multicolumn{1}{c}{Raw} & \multicolumn{1}{c}{DE} & \multicolumn{1}{c|}{PSD} & \multicolumn{1}{c}{Raw} & \multicolumn{1}{c}{DE} & \multicolumn{1}{c|}{PSD} & \multicolumn{1}{c}{Raw} & \multicolumn{1}{c}{DE} & \multicolumn{1}{c|}{PSD} & \multicolumn{1}{c}{Raw} & \multicolumn{1}{c}{DE} & \multicolumn{1}{c}{PSD} \\
CG & 69.862 & {\ul 74.066} & 68.958 & 66.496 & {\ul 75.758} & \textbf{73.386} & 72.641 & \textbf{66.397} & {\ul \textbf{73.218}} & 72.091 & 73.629 & {\ul 78.281} & 70.948 & {\ul \textbf{72.113}} & 65.179 \\
PCC & 66.474 & {\ul 72.281} & 72.199 & 68.735 & {\ul 75.819} & 73.165 & {\ul \textbf{74.001}} & 63.749 & 72.422 & 75.768 & \textbf{74.842} & {\ul 78.890} & {\ul 73.145} & 71.002 & 65.195 \\
MI & 68.874 & {\ul \textbf{74.242}} & 70.553 & 69.865 & {\ul 75.197} & 72.498 & {\ul 72.285} & 64.058 & 72.124 & \textbf{76.534} & 72.094 & {\ul 77.516} & {\ul \textbf{73.942}} & 69.999 & 65.770 \\
AEL & \textbf{71.160} & 72.458 & {\ul \textbf{73.873}} & \textbf{70.000} & {\ul \textbf{77.524}} & 72.490 & {\ul 72.488} & 65.633 & 72.008 & 72.902 & 73.281 & {\ul \textbf{79.327}} & 70.684 & 71.965 & {\ul \textbf{71.967}} \\ \hline
\end{tabular}
}
\end{table*}

\begin{table*}[]
\caption{Average test accuracy (\%) on ECG dataset type achieved by the benchmarked approaches and tested on five graph neural networks. Bold values indicate the edge feature learning method that performs best for a specific node feature, while underlined values indicate the node feature extraction method that performs best for a specific edge feature learning method.}
\label{table:ECG}
\resizebox{1.0\linewidth}{!}{
\begin{tabular}{llll|lll|lll|lll|lll}
\hline
 & \multicolumn{3}{c|}{\textbf{ChebNet}} & \multicolumn{3}{c|}{\textbf{GCN}} & \multicolumn{3}{c|}{\textbf{GAT}} & \multicolumn{3}{c|}{\textbf{MEGAT}} & \multicolumn{3}{c}{\textbf{STGCN}} \\ \hline
 & \multicolumn{1}{c}{Raw} & \multicolumn{1}{c}{DE} & \multicolumn{1}{c|}{PSD} & \multicolumn{1}{c}{Raw} & \multicolumn{1}{c}{DE} & \multicolumn{1}{c|}{PSD} & \multicolumn{1}{c}{Raw} & \multicolumn{1}{c}{DE} & \multicolumn{1}{c|}{PSD} & \multicolumn{1}{c}{Raw} & \multicolumn{1}{c}{DE} & \multicolumn{1}{c|}{PSD} & \multicolumn{1}{c}{Raw} & \multicolumn{1}{c}{DE} & \multicolumn{1}{c}{PSD} \\
CG & {\ul 56.666} & \textbf{53.334} & 46.666 & {\ul 53.334} & 53.333 & 43.334 & {\ul 60.000} & 56.666 & 40.000 & {\ul \textbf{60.000}} & \textbf{56.666} & 46.667 & {\ul 53.333} & 43.334 & 33.333 \\
PCC & {\ul 56.666} & 50.000 & 40.000 & {\ul 56.666} & {\ul \textbf{56.666}} & \textbf{46.667} & {\ul 63.333} & 53.334 & 40.000 & {\ul 56.666} & {\ul \textbf{56.666}} & 46.667 & {\ul 53.333} & 50.000 & \textbf{40.000} \\
MI & {\ul \textbf{63.334}} & 50.000 & 46.666 & {\ul 56.666} & 53.334 & \textbf{46.667} & {\ul \textbf{63.334}} & 56.666 & 43.333 & {\ul 56.666} & {\ul \textbf{56.666}} & \textbf{50.000} & \textbf{56.666} & {\ul \textbf{60.000}} & 33.333 \\
AEL & 46.666 & 50.000 & {\ul \textbf{60.000}} & {\ul \textbf{63.334}} & \textbf{56.666} & 43.334 & 50.000 & {\ul \textbf{60.000}} & \textbf{56.666} & {\ul 50.000} & {\ul 50.000} & {\ul \textbf{50.000}} & {\ul 50.000} & 46.666 & 36.666 \\ \hline
\end{tabular}
}
\end{table*}

\begin{table*}[]
\caption{Average test accuracy (\%) on EEG dataset type achieved by the benchmarked approaches and tested on five graph neural networks. Bold values indicate the edge feature learning method that performs best for a specific node feature, while underlined values indicate the node feature extraction method that performs best for a specific edge feature learning method.}
\label{table:EEG}
\resizebox{1.0\linewidth}{!}{
\begin{tabular}{llll|lll|lll|lll|lll}
\hline
 & \multicolumn{3}{c|}{\textbf{ChebNet}} & \multicolumn{3}{c|}{\textbf{GCN}} & \multicolumn{3}{c|}{\textbf{GAT}} & \multicolumn{3}{c|}{\textbf{MEGAT}} & \multicolumn{3}{c}{\textbf{STGCN}} \\ \hline
 & \multicolumn{1}{c}{Raw} & \multicolumn{1}{c}{DE} & \multicolumn{1}{c|}{PSD} & \multicolumn{1}{c}{Raw} & \multicolumn{1}{c}{DE} & \multicolumn{1}{c|}{PSD} & \multicolumn{1}{c}{Raw} & \multicolumn{1}{c}{DE} & \multicolumn{1}{c|}{PSD} & \multicolumn{1}{c}{Raw} & \multicolumn{1}{c}{DE} & \multicolumn{1}{c|}{PSD} & \multicolumn{1}{c}{Raw} & \multicolumn{1}{c}{DE} & \multicolumn{1}{c}{PSD} \\
CG & {\ul 65.316} & 58.177 & 54.246 & {\ul 61.955} & \textbf{59.859} & 60.094 & {\ul 64.585} & 56.516 & \textbf{58.283} & {\ul 66.541} & 56.862 & \textbf{59.571} & {\ul 66.644} & 57.722 & 59.016 \\
PCC & {\ul \textbf{65.594}} & 58.602 & \textbf{56.358} & {\ul 62.850} & 57.144 & \textbf{61.307} & {\ul 64.439} & 56.591 & 55.390 & {\ul 67.002} & 57.416 & 57.635 & {\ul 66.900} & 58.107 & \textbf{60.459} \\
MI & {\ul 62.458} & 58.267 & 55.460 & {\ul \textbf{65.112}} & 57.886 & 60.184 & {\ul 65.040} & 56.599 & 55.885 & {\ul 66.176} & 57.123 & 58.164 & {\ul \textbf{66.942}} & 58.517 & 59.778 \\
AEL & {\ul 63.206} & \textbf{59.286} & 54.282 & {\ul 62.874} & 59.273 & 60.723 & {\ul \textbf{65.996}} & \textbf{58.096} & 55.861 & {\ul \textbf{67.062}} & \textbf{60.101} & 58.858 & {\ul 63.705} & \textbf{59.513} & 60.292 \\ \hline
\end{tabular}
}
\end{table*}

\textbf{Comparison between edge feature learning:} As shown in Tables \ref{table:resultwithfs} and \ref{table:resultwithoutfs}, no single edge feature learning method consistently outperforms others across all settings. The optimal method depends on the combination of node features, GNN architecture, and the dataset. Even with the same node features and GNN model, the best-performing edge feature method can vary. For instance, with Raw series as node features and the GAT model in Table \ref{table:resultwithfs}, PCC achieves the best results. However, in Table \ref{table:resultwithoutfs}, AEL outperforms PCC with the same pairing. Overall, AEL demonstrates greater robustness, achieving the best performance in 7 out of 15 pairings in Table \ref{table:resultwithfs} and 3 out of 5 pairings in Table \ref{table:resultwithoutfs}, accounting for approximately half of the cases. This suggests that automatically learned edge features (via AEL) adapt better to diverse datasets, providing a more flexible solution.

\textbf{Comparison between single-dimensional and multi-dimensional edges:} As shown in Tables \ref{table:resultwithfs} and \ref{table:resultwithoutfs}, MEGAT consistently outperforms GAT, especially when DE and PSD are used as node features. For instance, MEGAT achieves an average improvement of 6\% in accuracy over GAT on DE and PSD across the datasets in Table \ref{table:resultwithfs}. This demonstrates the effectiveness of multi-dimensional edge representations in capturing complex relationships between nodes, enabling better node-to-node information exchange, even when node feature information is limited.

\begin{figure}[H]
    \centering
    \subfigure[Raw]{
        \includegraphics[width = 0.14\textwidth]{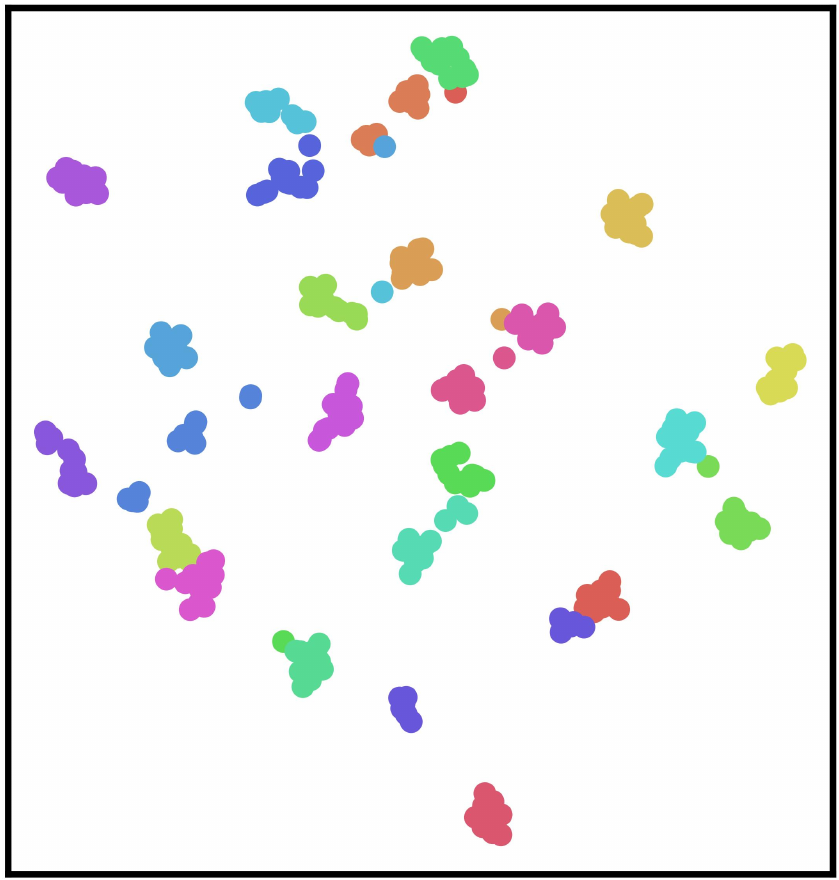}
    }
    \subfigure[DE]{
    \includegraphics[width = 0.14\textwidth]{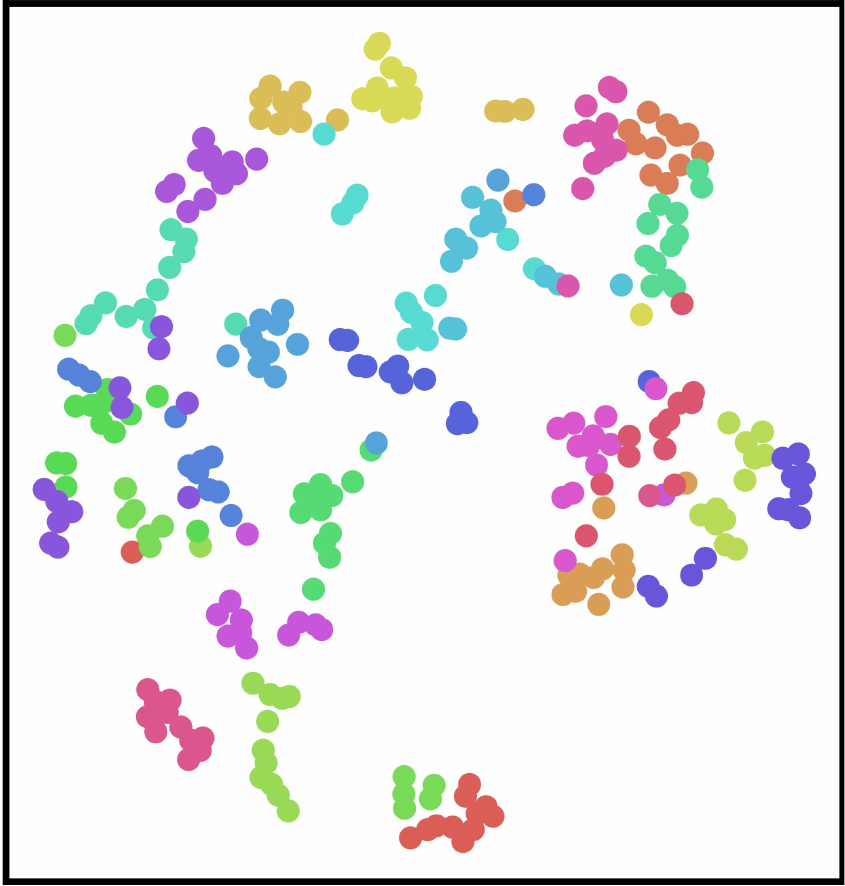}
    }
    \subfigure[PSD]{
    \includegraphics[width = 0.14\textwidth]{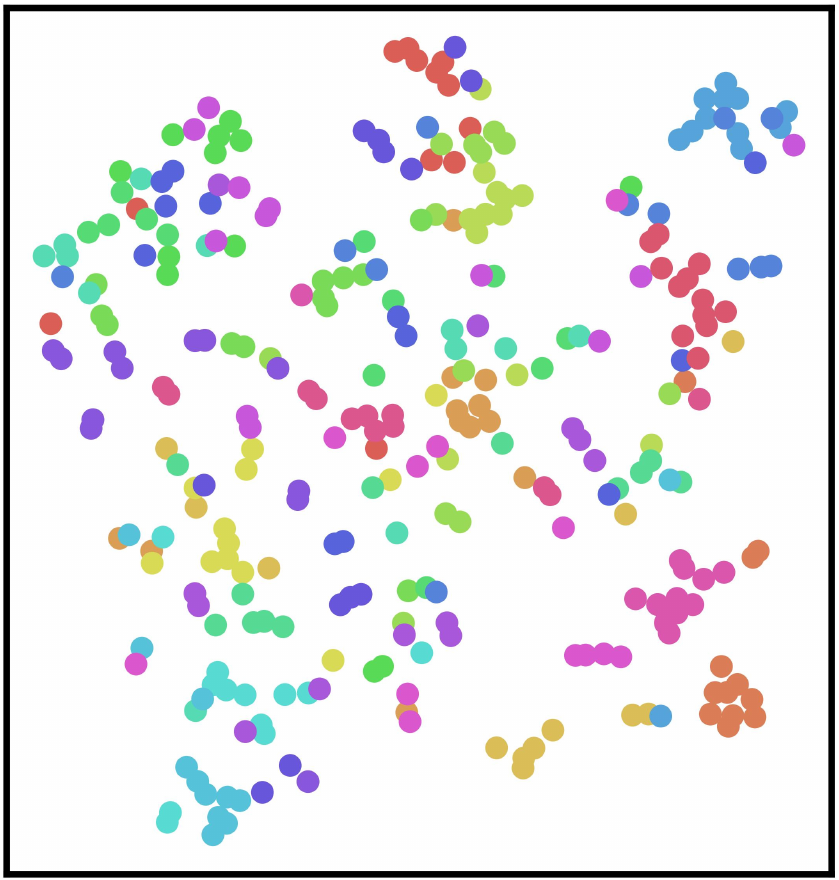}
    }
    \caption{Feature distribution of different node features of ArticularyWordRecognition dataset. The subfigure (a) shows the raw series and the subfigure, (b) presents the raw series after DE, (c) presents the raw series after PSD.}
    \label{fig:art-tsn}

\end{figure}

\begin{figure}[H]
    \centering
    \subfigure[Raw]{
        \includegraphics[width = 0.14\textwidth]{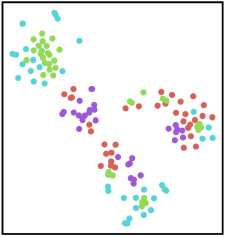}
    }
    \subfigure[DE]{
    \includegraphics[width = 0.14\textwidth]{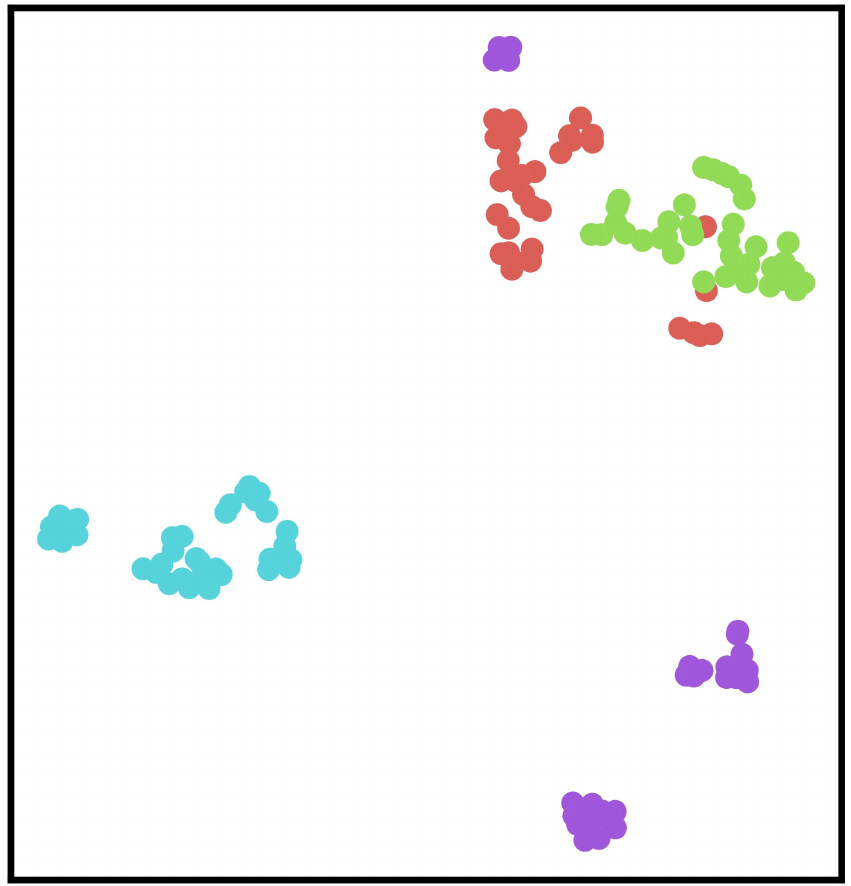}
    }
    \subfigure[PSD]{
    \includegraphics[width = 0.14\textwidth]{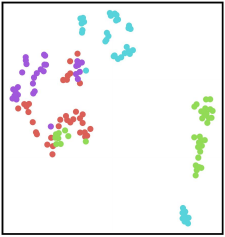}
    }
    \caption{Feature distribution of different node features of Epilepsy dataset. The subfigure (a) shows the raw series and the subfigure, (b) presents the raw series after DE, (c) presents the raw series after PSD.}
    \label{fig:epile-tsn}

\end{figure}

\textbf{Comparison between GNN architectures:} Tables \ref{table:resultwithfs} and \ref{table:resultwithoutfs} demonstrate that MEGAT consistently outperforms other networks, primarily due to its ability to leverage multi-dimensional edge features. In contrast, other models show inconsistent advantages, with their performance varying depending on the combinations of node features, edge features, and datasets. However, the impact of model selection on performance is less pronounced than the influence of input data quality. Specifically, the richness of node features and the accuracy of edge representations play a more critical role in determining final performance than the choice of GNN architecture. This highlights the importance of optimizing input data quality to achieve the best results.

\begin{figure}[htbp]
    \centering
    \subfigure[CG]{
    \includegraphics[width = 0.2\textwidth]{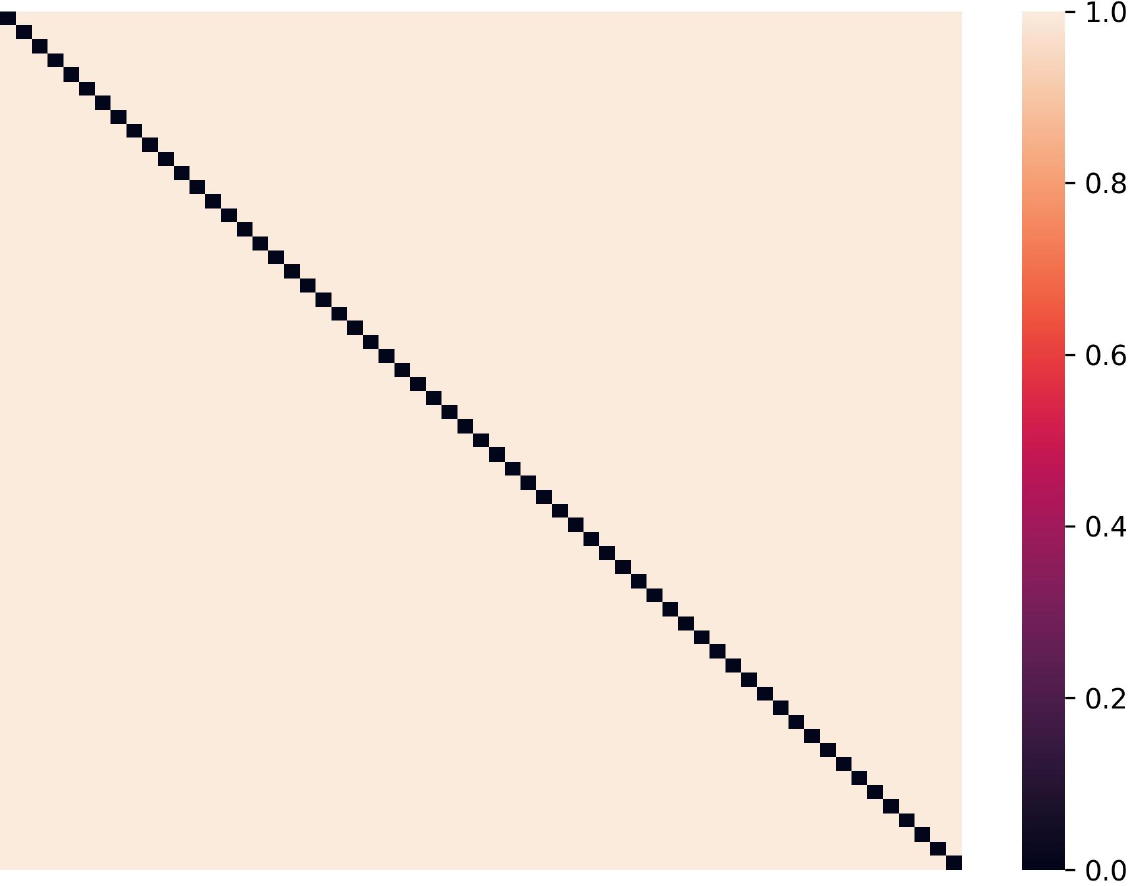}
    }
    \subfigure[PCC]{
        \includegraphics[width = 0.2\textwidth]{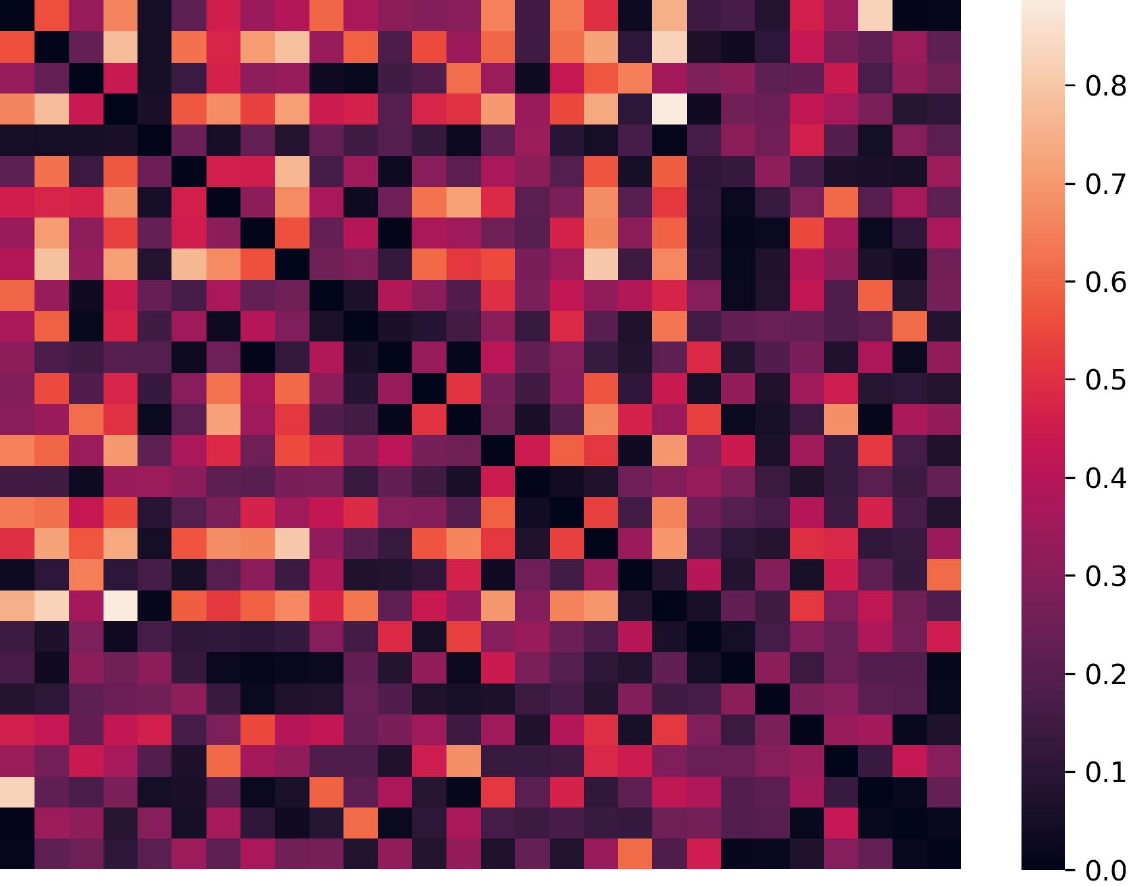}
    }
    \subfigure[MI]{
    \includegraphics[width = 0.2\textwidth]{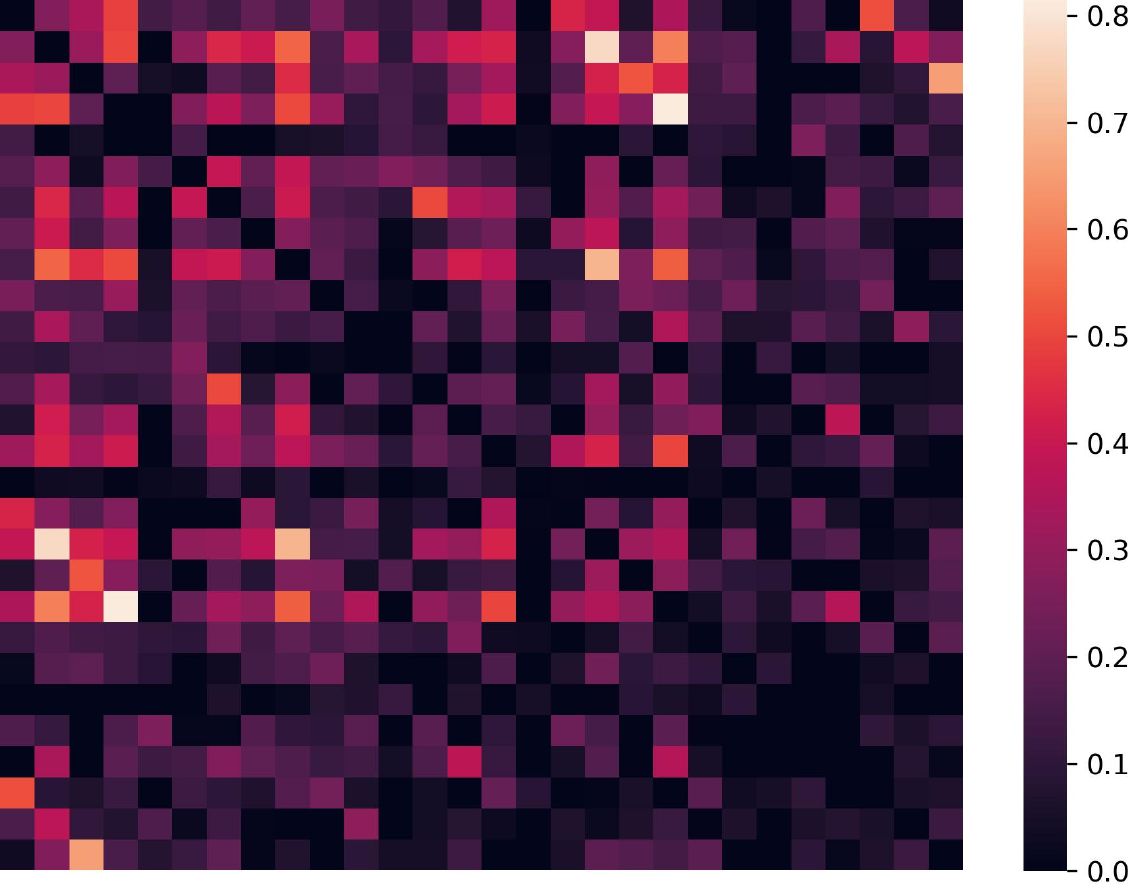}
    }
    \subfigure[AEL]{
    \includegraphics[width = 0.2\textwidth]{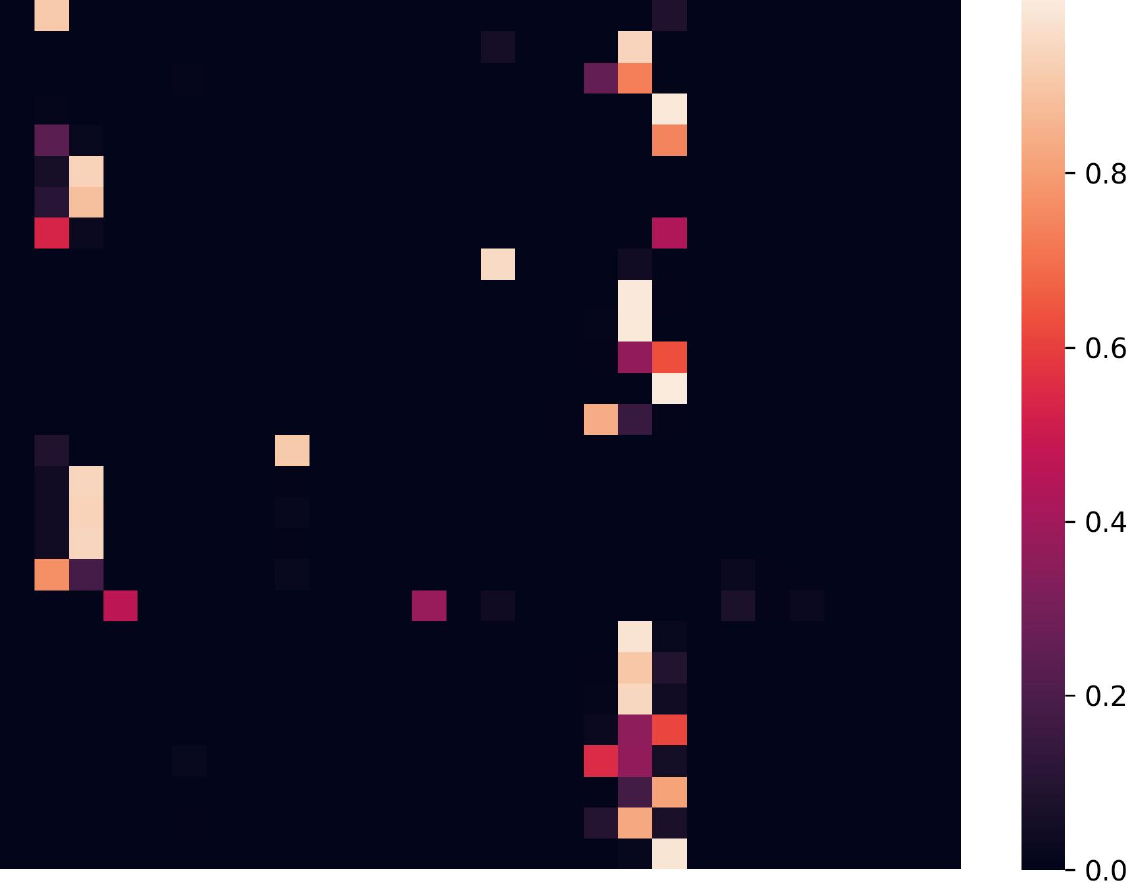}
    }
    \caption{Edge feature visualisation results for a sample in the FingerMovements dataset}
    \label{fig:finger-edge}

\end{figure}

\begin{figure}[htbp]
    \centering
    \subfigure[CG]{
    \includegraphics[width = 0.2\textwidth]{figure/complete.pdf}
    }
    \subfigure[PCC]{
        \includegraphics[width = 0.2\textwidth]{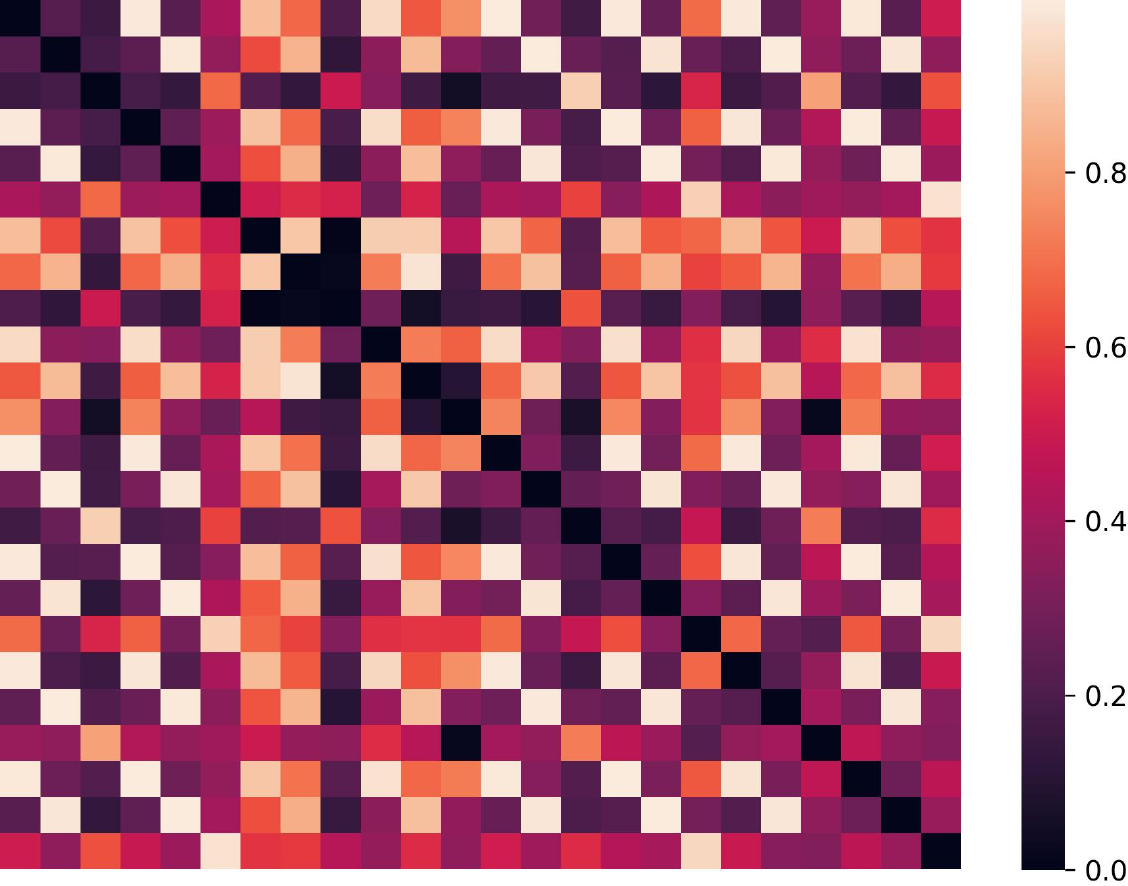}
    }
    \subfigure[MI]{
    \includegraphics[width = 0.2\textwidth]{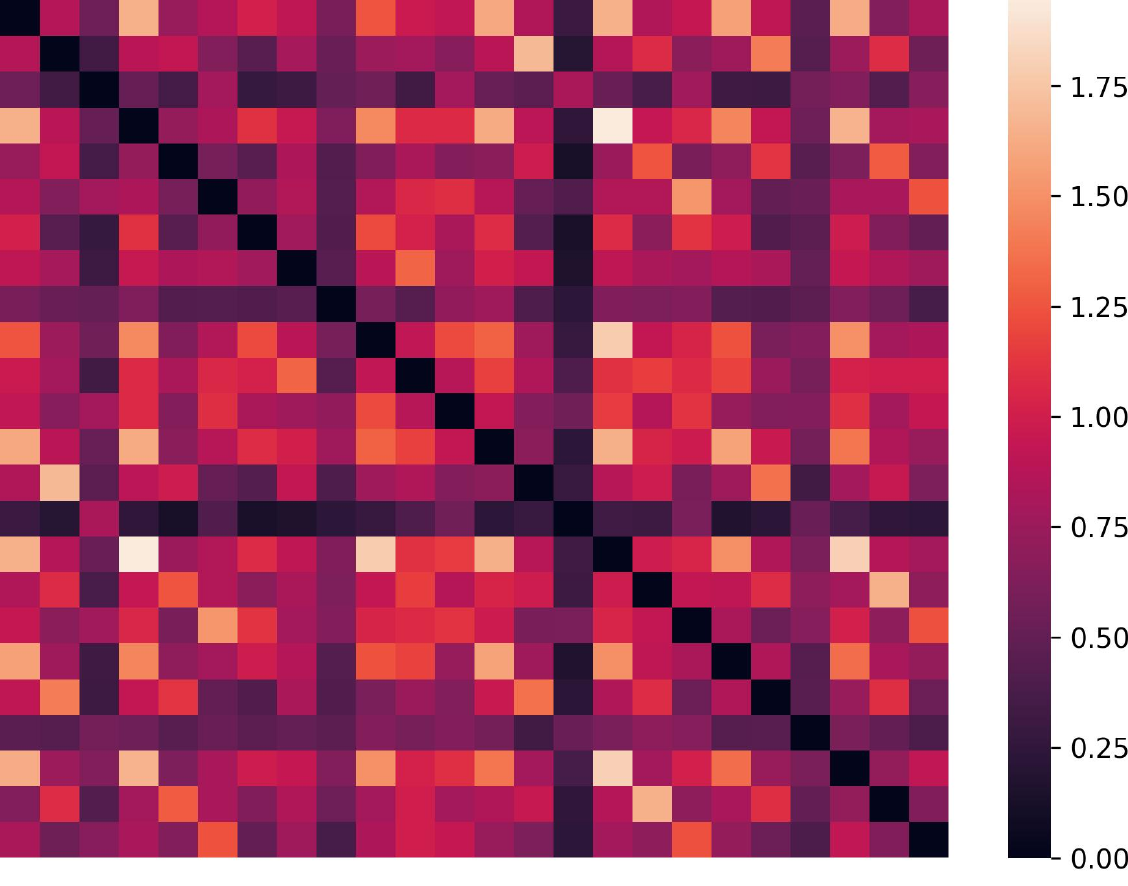}
    }
    \subfigure[AEL]{
    \includegraphics[width = 0.2\textwidth]{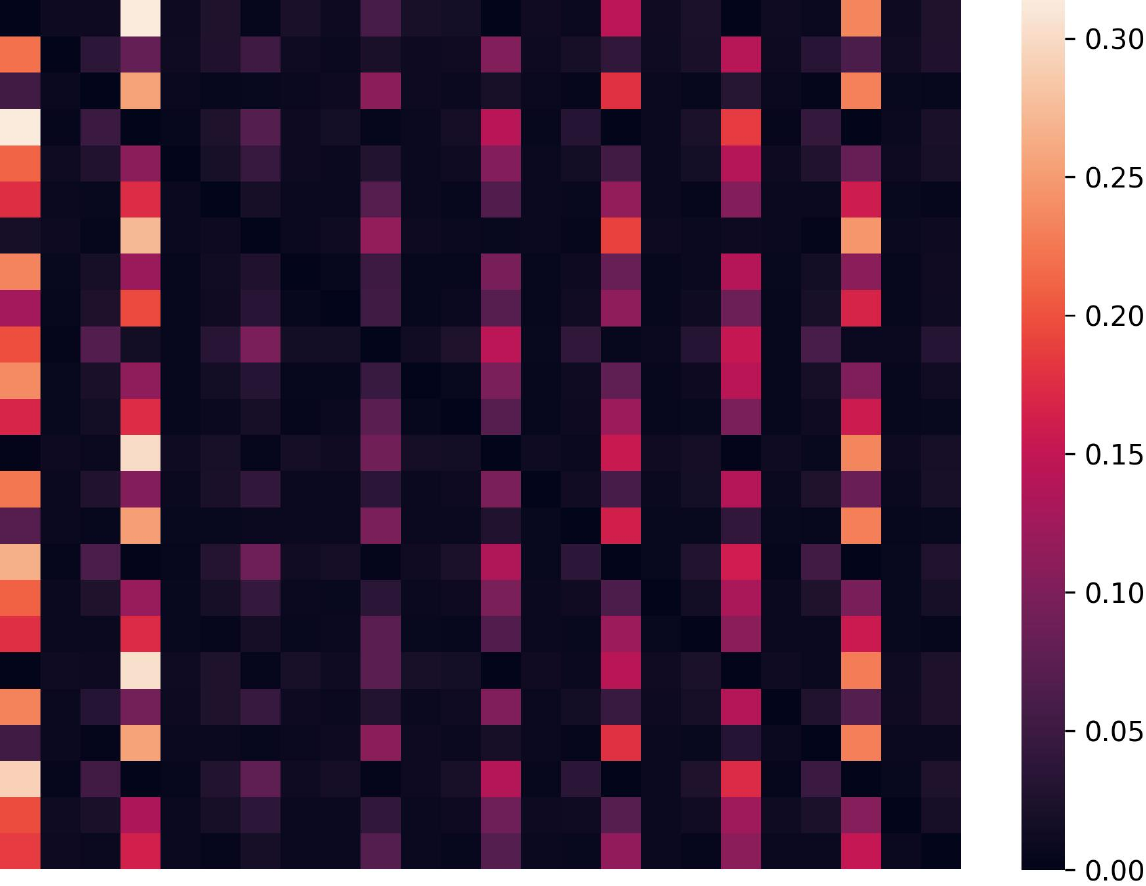}
    }
    \caption{Edge feature visualisation results for a sample in the NATOPS dataset}
    \label{fig:motor-edge}

\end{figure}

\begin{table*}[t]
\caption{Average test accuracy (\%) on ArticularyWordRecognition dataset achieved by the benchmarked approaches and tested on five graph neural networks.}
\label{table:ArticularyWordRecognition}
\resizebox{1.0\linewidth}{!}{
\begin{tabular}{llll|lll|lll|lll|lll}
\hline
 & \multicolumn{3}{c|}{\textbf{ChebNet}} & \multicolumn{3}{c|}{\textbf{GCN}} & \multicolumn{3}{c|}{\textbf{GAT}} & \multicolumn{3}{c|}{\textbf{MEGAT}} & \multicolumn{3}{c}{\textbf{STGCN}} \\ \hline
 & \multicolumn{1}{c}{Raw} & \multicolumn{1}{c}{DE} & \multicolumn{1}{c|}{PSD} & \multicolumn{1}{c}{Raw} & \multicolumn{1}{c}{DE} & \multicolumn{1}{c|}{PSD} & \multicolumn{1}{c}{Raw} & \multicolumn{1}{c}{DE} & \multicolumn{1}{c|}{PSD} & \multicolumn{1}{c}{Raw} & \multicolumn{1}{c}{DE} & \multicolumn{1}{c|}{PSD} & \multicolumn{1}{c}{Raw} & \multicolumn{1}{c}{DE} & \multicolumn{1}{c}{PSD} \\
CG & 74.333 & 61.667 & 39.667 & 67.000 & 42.333 & 23.000 & 82.667 & 34.333 & 9.000 & 72.000 & 59.333 & 30.333 & 65.667 & 43.667 & 7.333 \\
PCC & 79.333 & 63.333 & 40.000 & 69.667 & 45.000 & 23.333 & 79.000 & 34.000 & 9.000 & 76.000 & 59.333 & 35.667 & 66.667 & 50.333 & 7.000 \\
MI & 82.667 & 51.333 & 41.667 & 69.667 & 46.000 & 23.667 & 82.000 & 34.333 & 9.000 & 74.000 & 61.000 & 32.667 & 70.667 & 53.333 & 7.333 \\
AEL & 79.667 & 56.333 & 46.000 & 77.667 & 66.333 & 28.000 & 81.667 & 29.667 & 10.000 & 71.000 & 63.667 & 28.000 & 67.000 & 51.667 & 5.333 \\ \hline
\end{tabular}
}
\end{table*}

\begin{table*}[htbp]
\caption{Average test accuracy (\%) on Epilepsy dataset achieved by the benchmarked approaches and tested on five graph neural networks.}
\label{table:Epilepsy}
\resizebox{1.0\linewidth}{!}{
\begin{tabular}{llll|lll|lll|lll|lll}
\hline
 & \multicolumn{3}{c|}{\textbf{ChebNet}} & \multicolumn{3}{c|}{\textbf{GCN}} & \multicolumn{3}{c|}{\textbf{GAT}} & \multicolumn{3}{c|}{\textbf{MEGAT}} & \multicolumn{3}{c}{\textbf{STGCN}} \\ \hline
 & \multicolumn{1}{c}{Raw} & \multicolumn{1}{c}{DE} & \multicolumn{1}{c|}{PSD} & \multicolumn{1}{c}{Raw} & \multicolumn{1}{c}{DE} & \multicolumn{1}{c|}{PSD} & \multicolumn{1}{c}{Raw} & \multicolumn{1}{c}{DE} & \multicolumn{1}{c|}{PSD} & \multicolumn{1}{c}{Raw} & \multicolumn{1}{c}{DE} & \multicolumn{1}{c|}{PSD} & \multicolumn{1}{c}{Raw} & \multicolumn{1}{c}{DE} & \multicolumn{1}{c}{PSD} \\
CG & 63.043 & 99.275 & 91.304 & 51.449 & 99.275 & 91.304 & 54.348 & 97.101 & 91.304 & 56.522 & 97.101 & 92.029 & 55.072 & 99.275 & 92.029 \\
PCC & 57.246 & 99.275 & 89.130 & 54.348 & 99.275 & 89.855 & 55.797 & 97.826 & 89.855 & 65.217 & 98.551 & 94.203 & 64.493 & 99.275 & 94.928 \\
MI & 54.348 & 99.275 & 88.406 & 55.797 & 99.275 & 90.580 & 52.174 & 97.101 & 89.855 & 68.116 & 99.275 & 93.478 & 71.014 & 99.275 & 92.754 \\
AEL & 63.768 & 97.826 & 91.304 & 54.348 & 99.275 & 92.754 & 55.072 & 95.652 & 89.855 & 56.522 & 97.101 & 93.478 & 55.797 & 99.275 & 93.478 \\ \hline
\end{tabular}
}
\end{table*}

\begin{figure*}[]
    \centering
\includegraphics[width=1.0\linewidth]{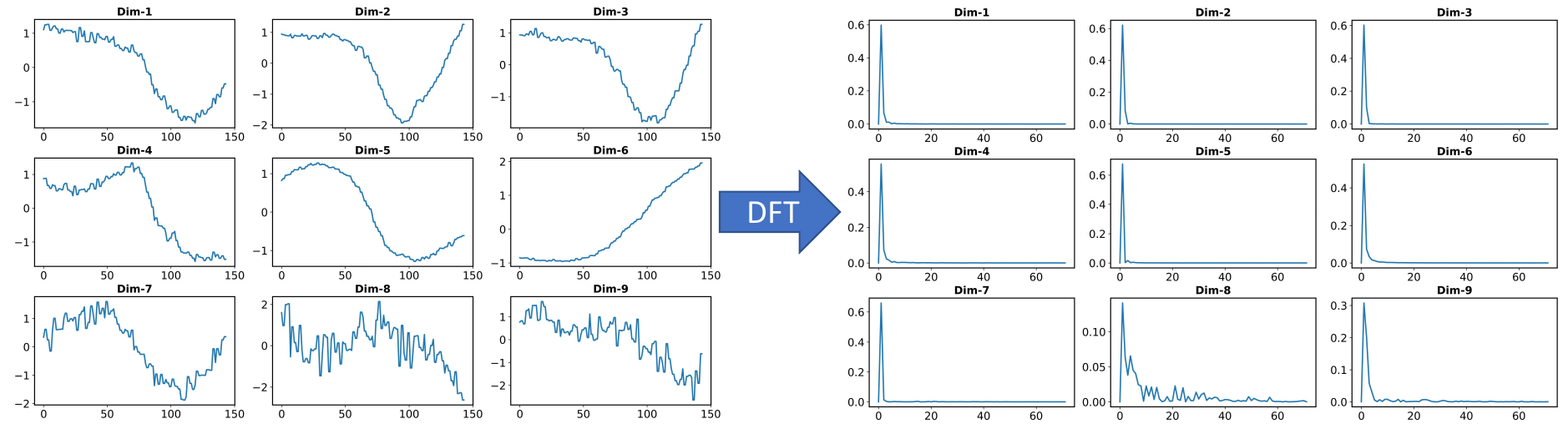}
    \caption{Raw series and spectrum of the ArticularyWordRecognition dataset}
    \label{fig:art}
\end{figure*}

\begin{figure*}[]
    \centering
\includegraphics[width=1.0\linewidth]{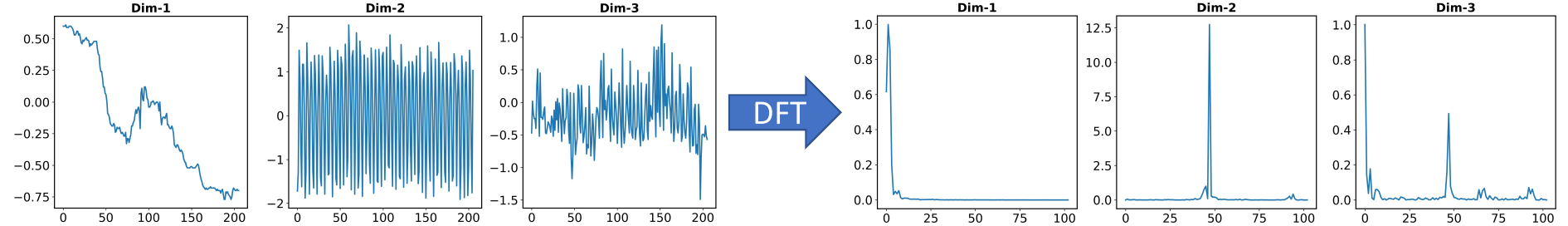}
    \caption{Raw series and spectrum of the Epilepsy dataset.}
    \label{fig:epil}
\end{figure*}

\subsection{Analysis and discussion}
\label{sec:analysis}
In this section, we evaluate the performance of the benchmarked approaches across various dataset types, with a focus on how well each method aligns with specific application. We provide actionable insights for selecting node features by analyzing dataset properties, including data distribution. Furthermore, we visualize edge features to identify meaningful patterns and gain deeper insights into the observed performance differences.

\textbf{Comparison of different dataset types:} We categorized the datasets in Table \ref{table:resultwithfs} into three types: HAR (Table \ref{table:HAR}), ECG (Table \ref{table:ECG}) and EEG (Table \ref{table:EEG}). On HAR datasets, DE and PSD demonstrate strong performance, often surpassing raw series as node features. This suggests that frequency-domain features are particularly effective in capturing the patterns present in HAR data, which we explore further in the following. In contrast, MEGAT consistently outperforms GAT across most cases, regardless of the dataset type. This highlights the advantages of leveraging multi-dimensional edge features, which enhance the model's ability to capture complex relationships between data regardless of data type.

\textbf{How to choose node feature extraction method:} To explore the impact of node feature extraction methods and determine the most suitable approach for specific datasets, we analyzed two datasets where node features significantly influenced classification results. For the \textbf{ArticularyWordRecognition} dataset, Table \ref{table:ArticularyWordRecognition} highlights substantial performance differences across feature extraction methods, consistent across various networks. As shown in Fig. \ref{fig:art-tsn}, the original data exhibits a coherent and compact feature distribution, whereas DE and PSD introduce greater dispersion. This increased dispersion correlates with performance degradation, with PSD showing more pronounced dispersion and worse classification results than DE. These findings suggest that feature dispersion negatively impacts model performance, particularly when feature distributions become overly scattered. In contrast, for the \textbf{Epilepsy} dataset (a HAR dataset), DE and PSD improve performance, as indicated in Table \ref{table:Epilepsy}. Visualization in Fig. \ref{fig:epile-tsn} shows that these methods produce more aggregated feature distributions within the same category, resulting in enhanced classification accuracy. This emphasizes that aggregated feature distributions are more conducive to model performance, highlighting the utility of visualizing feature distributions when selecting node features. Furthermore, both DE and PSD operate in the frequency domain, where spectrum visualizations offer valuable insights (Fig. \ref{fig:art} and \ref{fig:epil}). For the ArticularyWordRecognition dataset, frequency domain conversion reduces inter-variable differences, diminishing the information content of the series and resulting in poorer performance. Conversely, for the Epilepsy dataset, frequency domain conversion emphasizes inter-variable distinctions, effectively capturing critical information and boosting performance.

\textbf{Visualisation of edge features:} To analyze the characteristics of edge features, we randomly selected two samples for visualization. To highlight inter-node relationship strengths, we set self-loop weights to 0. Fig. \ref{fig:finger-edge} and \ref{fig:motor-edge} illustrate the edge weight distributions for CG, PCC, MI, and AEL. The distributions for PCC and MI are structurally similar, with differences primarily in the absolute weight values. This suggests that both methods capture similar patterns of inter-node relationships. In contrast, AEL generates highly concentrated edge weights, where specific nodes dominate the learned relationships while most edges are assigned near-zero weights. This sparsity reflects AEL`s ability to selectively emphasize critical interactions, reducing noise from less relevant connections. By focusing on key relationships, AEL potentially enhances the model’s capacity to capture meaningful structural information, which is especially valuable in complex datasets where irrelevant connections might hinder performance.

\section{Conclusion}
In this study, we introduce a benchmarking framework designed to evaluate the efficacy of various graph representation learning methods and graph networks in the context of multivariate time series classification. The code for this framework will be made publicly available to support future research in this area. Our benchmark involved three node features, four edge features, four graph networks and a GAT network with the addition of multidimensional edge feature learning (MEGAT).

The experimental results highlight three key findings:
\begin{itemize}
    \item Impact of Node Features: Node features play a crucial role in classification accuracy. To select suitable node features, we recommend comparing raw and spectralized features or analyzing feature distributions. Methods that enhance variable differentiation or produce more clustered feature distributions within the same category generally yielded better results.
    
    \item Edge Feature Visualization: Visualization of edge features demonstrates that adaptive edge learning effectively emphasizes key node-to-node relationships while minimizing the influence of irrelevant ones. This approach achieves the best classification performance in over 50\% of cases, showcasing its robustness and adaptability. 
    
    \item Effectiveness of Multidimensional Edge Features: Multidimensional edge feature graph attention network significantly improve graph attention network's classification accuracy across diverse datasets. This finding highlights its ability to capture complex inter-node relationships, underscoring its importance in graph representation learning.
\end{itemize}
While this study does not integrate graph pooling techniques, our framework is designed to seamlessly accommodate their incorporation. Future work will explore the potential of graph pooling in enhancing multivariate time series classification, aiming to better capture graph structures and improve model performance.

%
\bibliographystyle{IEEEtran}
\bibliography{reference}

\end{document}